\documentclass[11pt]{article}

\usepackage[preprint]{acl}

\usepackage{times}
\usepackage{latexsym}

\usepackage{amsmath}

\usepackage[T1]{fontenc}
\usepackage[utf8]{inputenc}

\usepackage{microtype}

\usepackage{inconsolata}

\usepackage{graphicx}
\usepackage{listings}
\usepackage{xspace}
\usepackage{multirow}
\usepackage{booktabs}
\usepackage{xcolor}
\usepackage{realboxes}
\usepackage{amssymb}
\usepackage{pifont}
\usepackage{tabularx}
\usepackage[most]{tcolorbox}
\usepackage{enumitem}
\usepackage{makecell}
\tcbset{
promptbox/.style={
colback=gray!8,
colframe=black,
colbacktitle=black,
coltitle=white,
boxsep=1mm,
left=2mm,
right=2mm,
breakable,
fonttitle=\bfseries\small,
fontupper=\small,
}
}

\definecolor{darkgreen}{rgb}{0.0,0.5,0.0}
\definecolor{darkred}{rgb}{0.7,0.0,0.0}
\definecolor{darkbrown}{rgb}{0.6, 0.2, 0.0}
\definecolor{toolbg}{HTML}{F2F2F2} % Light gray background

\newcommand{\benchmark}{\textsc{AgenticInterpBench}\xspace}
\newcommand{\agent}{\textsc{HyVE}\xspace}
\newcommand{\tool}[1]{\colorbox{toolbg}{\texttt{#1}}}

\title{Can Language Model Agents be Helpful Circuit Explainers in Mechanistic Interpretability?}
\author{
\textbf{Ayan Antik Khan\textsuperscript{1}},
\textbf{Harsh Kohli\textsuperscript{2}},
\textbf{Yuekun Yao\textsuperscript{2}},
\\
\textbf{Huan Sun\textsuperscript{2}},
\textbf{Ziyu Yao\textsuperscript{1}}
\\
\\
\textsuperscript{1}George Mason University
\qquad
\textsuperscript{2}The Ohio State University
\\
\small{
\texttt{\{akhan265,ziyuyao\}@gmu.edu}
\qquad
\texttt{\{kohli.120,yao.1267,sun.397\}@osu.edu}
}
}

\begin{document}
\maketitle
\begin{abstract}
Mechanistic interpretability has made substantial progress in automatically localizing circuits, but explaining what localized components do remains labor-intensive and difficult to standardize. In this work, we study whether language model (LM) agents can assist with this explanation problem once a circuit has already been identified. We introduce \textbf{\benchmark}, a benchmark for circuit explanation built from 84 semi-synthetic transformer circuits with 163 component-level annotations. We propose \textbf{\agent} (\underline{\textbf{Hy}}pothesize, \underline{\textbf{V}}alidate, \underline{\textbf{E}}xplain), an agentic explainer that analyzes each component through an iterative loop of observation, hypothesis generation, and causal validation, eventually producing a component-level explanation and a circuit-level task description. 
Across four LM backbones, \agent recovers useful component- and task-level explanations, but no backbone is uniformly best. Our analysis shows that strong backbones usually form observation-grounded hypotheses, while failures more often arise later in the validation loop, through incomplete validation plans, code execution errors, or unresolved hypotheses.
A case study on an arithmetic circuit in Llama-3-8B shows that the same formulation can extend beyond semi-synthetic benchmarks to naturally trained models. Overall, LM agents are promising circuit explainers, but reliable validation remains the key obstacle.\footnote{We release the benchmark dataset, source code, and prompts at \url{https://github.com/Ziyu-Yao-NLP-Lab/LLM-Circuit-Explainer}.}
\end{abstract}

\section{Introduction}\label{sec:intro}

\begin{figure}[t]
    \centering
    \includegraphics[width=\linewidth]{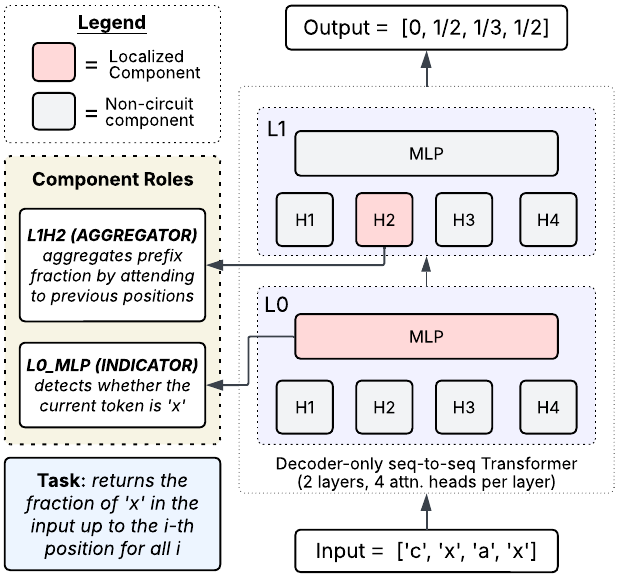}
    \caption{An instance of the circuit explanation task on \texttt{frac\_prevs}, a 2-layer transformer that computes the running fraction of token `x'. An agent receives the input-output examples and a localized circuit and must (i) assign each component a functional role (e.g., L0\_MLP: INDICATOR, L1H2: AGGREGATOR) along with a natural-language role description and (ii) derive a task description of the overall model behavior.
    }
    \label{fig:io}
\end{figure}

Mechanistic Interpretability (MI) seeks to reverse-engineer how language models (LMs) implement specific behaviors by identifying the underlying \textit{circuits}, i.e., sub-networks of attention heads and MLP components that contribute to the model behaviors \citep{rai2024practical, bereska2024mechanistic, ferrando2024primerinnerworkingstransformerbased}. While recent advances have made circuit \textit{localization} more efficient through automated patching and attribution methods \cite{conmy2023towards, hanna2024faithfaithfulnessgoingcircuit, syed2024attribution}, the \textit{explanation} phase, i.e., understanding the semantic roles of these components and their interactions, remains largely manual and difficult to scale.
Human researchers typically conduct iterative hypothesis generation and validation using established methods. Yet as models grow larger and more complex, this human-centered process becomes increasingly infeasible. {Recent work has shown that LM agents can support open-ended scientific workflows by generating hypotheses, designing experiments, executing code, and refining conclusions from evidence \citep{chen2025scienceagentbench, lu2024ai, yamada2025ai}. Given that circuit explanation shares a similar loop, a natural question arises: \textit{Can LM agents assist in explaining the circuits within an LM?}
\par In this work, we explore whether LM agents can work as effective circuit explainers once a circuit is localized. We focus on assessing the sufficiency and reliability of LMs in generating and validating explanations grounded in mechanistic evidence, an essential step toward scalable and automated circuit understanding. 
To study this problem in a controlled setting, we construct \benchmark, a benchmark comprising 84 localized circuits on semi-synthetic transformers that cover 163 transformer components, built on the InterpBench~\cite{gupta2025interpbench}.
Each component is annotated with a functional role tag drawn from a 5-class taxonomy together with a natural-language description of its task-specific role. 
{This semi-synthetic setting enables systematic evaluation with component-level ground truth and, importantly, reduces the risk of \emph{data leakage}. That is, on well-known natural circuits, agents may recall published component roles instead of deriving them through mechanistic experiments \citep{haklay2026pitfallsevaluatinginterpretabilityagents}, which not only overestimates agent performance but also obscures the true reasoning and coding capability of agents on this task.
% \benchmark is built on InterpBench~\cite{gupta2025interpbench}, a suite of semi-synthetic transformers with known ground-truth circuits.
\par We further propose \textbf{\agent} (\underline{\textbf{Hy}}pothesize, \underline{\textbf{V}}alidate, \underline{\textbf{E}}xplain), an agent-based framework that explains a localized circuit through iterative \textit{observation, hypothesis generation, and validation}. {We evaluate \agent on \benchmark using four frontier LMs as backbones: GPT-5.4~\citep{openai2026gpt54}, Claude-Sonnet-4.6~\citep{anthropic2026claudesonnet46}, Gemini-3.1-Pro~\citep{googledeepmind2026gemini31pro}, and Qwen-3-Coder-30B-A3B-Instruct~\citep{qwen3technicalreport}. \agent achieves up to 79\% component tag accuracy and 83\% task accuracy. The results show that LM agents can produce useful circuit explanations, but no backbone is uniformly best. Initial hypotheses are usually grounded for the stronger backbones, while the main failures arise later in the validation loop. GPT-5.4 produces the soundest validation plans, Claude-Sonnet-4.6 executes code most reliably, and Gemini-3.1-Pro achieves the strongest judged explanation scores. These trends suggest that hypothesis generation may not be the main bottleneck by itself, yet reliable circuit explanation also depends on validation design and code execution.}

\par To evaluate whether \agent generalizes beyond the semi-synthetic transformers of \benchmark, we conduct a case study on a realistic circuit for three-operand addition in Llama-3-8b \citep{mamidanna-etal-2025-one}. 
% We evaluate \agent on a 10-component localized circuit and compare its component descriptions against manually derived reference roles. 
{Our experiment shows that \agent can recover component roles in this setting: Claude-Sonnet-4.6 correctly explains 8 of 10 components, while GPT-5.4 gives 6 correct and 3 partially correct descriptions. Both models recover the main operand-transfer structure, while Claude also explains the causally redundant components. This case study complements \benchmark by testing \agent in a more realistic setting, where the localized circuit comes from a naturally trained next-token prediction model. It also highlights a practical role for agentic explainers as tools for stress-testing existing circuit analyses and probing for missed mechanisms.}

\section{Related Work}\label{sec:related-work}

\paragraph{Mechanistic Interpretability (MI)}
MI has largely advanced through detailed case studies that localize and explain circuits for specific model behaviors. A landmark example is the IOI circuit of \citet{wang2023interpretability}, a 26-head circuit for indirect-object identification in GPT-2 small.
A complementary line of work studies arithmetic and algorithmic circuits in LMs, including greater-than comparison \citep{hanna2023doesgpt2computegreaterthan} and helical number representations \citep{kantamneni2025languagemodelsusetrigonometry}. 
In our work, we evaluate the generalizability of \agent on the All-for-One (AF1) subgraph discovered by \citet{mamidanna-etal-2025-one} for mental math.

\paragraph{Automation in MI} Early analyses relied solely on human-designed interventions to identify relevant model components. ACDC \citep{conmy2023towards} automates part of this process by pruning a model's computational graph with intervention-based tests. EAP and EAP-IG \citep{syed2024attribution,hanna2024faithfaithfulnessgoingcircuit} further improve scalability by using attribution-based scores to identify important circuit edges. These methods increasingly automate \textit{localization}, but the subsequent \textit{explanation} step still largely requires human analysis. Our work was motivated by the need to fill this gap.

Similar to us, \citet{2024arXiv241013928P, pmlr-v235-shaham24a, liu2026neuronscopemultiagentframeworkexplaining, marinllobet2026automatedinterpretabilityfeaturediscovery} explore automated interpretability; however, they focus on explaining isolated features or neurons, while we target circuit explanation (i.e., explaining transformer components and how they connect to enable specific task performance). 
{\citet{bissell2025researchagent} use agents to conduct open-ended interpretability experiments, and present practical workflow lessons; however, they do not address automated circuit explanation. \citet{singh2026agenticimodels} use agents to \textit{design} interpretable tabular models, whereas we use an agent to \textit{explain} internal circuits of trained LMs.}
Finally, \citet{bai2026storyscienceexecutiongroundedevaluation} design agents to \emph{evaluate} MI findings against its underlying code, data, and evidence, while we create agents to \emph{perform} MI research from scratch.
% \akc{Citing \cite{haklay2026pitfallsevaluatinginterpretabilityagents}.}
{Concurrent with our work, \citet{haklay2026pitfallsevaluatinginterpretabilityagents} develop a similar hypothesize-validate agent for circuit components but focus on \emph{evaluation}, showing that comparisons against published expert explanations is vulnerable to memorization. Our RASP-derived ground truth avoids this issue in the controlled setting, while the concern persists for naturally trained circuits.}

\paragraph{Benchmarks in MI}

MIB \citep{pmlr-v267-mueller25a} evaluates circuit localization by reporting two metrics derived from the faithfulness of the circuit against the full model.
Tracr \citep{lindner2023tracrcompiledtransformerslaboratory} compiles RASP \citep{weiss2021thinking} programs into transformers with known internal structure, which can then serve as the ground-truth circuits, and TracrBench \citep{thurnherr2024tracrbench} scales this approach. InterpBench \citep{gupta2025interpbench} builds on this line by producing more realistic transformers with known circuits. These datasets, however, are all evaluating the \emph{localization} of circuits, yet benchmarking the \emph{explanation} of circuit components remains widely understudied.
In this line, FIND \citep{2023arXiv230903886S} evaluates open-ended descriptions of black-box functions, but it does not center MI circuits. Our work fills this gap by proposing the first benchmark for agentic circuit explanation. Our benchmark was built on top of InterpBench, as described in Section~\ref{sec:benchmark}.

\section{Benchmarking LLM Agents as Circuit Explainers}
\label{sec:benchmark}

In this section, we formulate the task of circuit explanation and describe \benchmark.

\subsection{Task Formulation}
Formally, let $\mathcal{E} = \{(x_k, y_k)\}_{k=1}^m$ denote a set of task input-output examples illustrating the model's behavior, and let $\mathcal{C} = \{c_1, c_2, \dots, c_n\}$ denote a localized circuit, where each $c_i$ is a circuit component such as an attention head or MLP sublayer. The agent's task is to explain the functional role of each component and the task-level behavior implemented by the circuit. 
The agent produces \textbf{three} outputs. For each circuit component $c_i$, it predicts (i) a role tag $t_i\in \mathcal{R}$ summarizing the component's abstract role, where $\mathcal{R}$ is the role taxonomy introduced in Section~\ref{par:Annotation}, and (ii) a natural-language note $n_i$ describing the task-specific behavior of $c_i$. 
For the full circuit, it also produces (iii) a derived task description $d$ characterizing the LM's underlying task. Figure~\ref{fig:io} illustrates these inputs and outputs on the running example \texttt{frac\_prevs}.

\subsection{\benchmark}
We introduce \benchmark, a benchmark for evaluating LLM agents on circuit explanation. \benchmark consists of 84 transformer circuits with 163 annotated components (Table~\ref{tab:benchmark_stats}). Notably, \benchmark targets a circuit explanation setting in which the localized circuit is given and the agent must recover the role of each component. \benchmark is built on InterpBench \cite{gupta2025interpbench}, which we briefly review before describing our annotation taxonomy and construction procedure.

\begin{table}[t!]
\centering
\small
\begin{tabular}{lr}
\toprule
\textbf{Statistic} & \textbf{Count} \\
\midrule
Benchmark tasks/circuits & 84 \\
Total MLP components & 120 \\
Total attention components & 43 \\
Avg./Min/Max \#of components per circuit & 1.94/1/10 \\
% Tasks solved by a single MLP component & 51 \\
% Circuits w/ only MLP component(s) & 56 \\
\midrule
\multicolumn{2}{c}{\textit{Role tag counts}} \\
\midrule
MAPPER & 72 \\
AGGREGATOR & 32 \\
COMBINER & 33 \\
ROUTER & 11 \\
INDICATOR & 15 \\
\bottomrule
\end{tabular}
\caption{Statistics of \benchmark.}
\label{tab:benchmark_stats}
\end{table}

\paragraph{Background.} InterpBench provides semi-synthetic transformers whose ground-truth circuits are known by design. It builds on {Tracr}~\cite{lindner2023tracrcompiledtransformerslaboratory}, a compiler that converts {RASP} programs~\cite{weiss2021thinking} into decoder-only transformers with fully transparent computational structure. To mitigate the unrealistic weight distributions in Tracr-compiled models, InterpBench retrains Tracr models with \textit{Strict Interchange Intervention Training} (SIIT), a procedure extended from IIT \cite{pmlr-v162-geiger22a} that aligns a low-level transformer with the Tracr-compiled circuit while penalizing contributions from non-circuit components. The resulting models exhibit weight distributions and activations close to those of naturally trained transformers, while preserving the same circuit components of their Tracr counterparts.
\par We use the 84 RASP-derived models in InterpBench as the foundation for \benchmark.\footnote{We exclude the two IOI tasks, as IOI is a widely studied circuit and the agent may rely on memorized conclusions instead of grounded execution \cite{haklay2026pitfallsevaluatinginterpretabilityagents}.} These tasks span small algorithmic behaviors, including counting, fraction computation, sorting, and matching. Two properties make them well-suited for evaluating LMs as circuit explainers: (i) the ground-truth circuit and per-component role are recoverable from the RASP source, enabling precise evaluation, and (ii) the diversity of tasks reduces the risk of the agent memorizing well-known circuits from prior literature.

\paragraph{Dataset Annotation}
\label{par:Annotation}

{We build \benchmark by extending InterpBench with a semantic annotation layer for circuit explanation. For each localized component in an InterpBench model, we inspect the corresponding RASP program and use {InterpBench's high-level/low-level correspondence map to trace the trained component back to the RASP variable it implements.} This allows us to assign precise task-specific roles to each component.
}

Specifically, each task in \benchmark is annotated with its task description, the original RASP program, five input-output examples with inputs sampled from the task's data distribution and outputs obtained by executing the RASP program, and per-component \textit{role annotations}. 

A {role annotation} consists of two fields: a \textbf{\textit{tag}}, drawn from the 5-class taxonomy
{(\textsc{Indicator}, \textsc{Aggregator}, \textsc{Router}, \textsc{Mapper}, and \textsc{Combiner}) detailed in Appendix~\ref{app:role-taxonomy}},
and a \textbf{\textit{note}}, a brief natural-language description of the component's task-specific role. Together, the two fields support evaluation at two granularities: whether the agent identifies the correct \textit{abstract role}, and whether it can describe that role accurately in the task context. 
{The taxonomy is specific to \benchmark and is not intended as a comprehensive taxonomy of transformer component functions. Its coarse abstraction enables deterministic and scalable evaluation of tag predictions, complementing the natural-language explanation evaluation}.
The annotation was manually performed and examined against the original RASP program-circuit mapping to ensure quality.

{An example is shown in Figure~\ref{fig:io}. We include its corresponding RASP program and other details in Appendix~\ref{app:annotation-example}.}

\subsection{Evaluation Metrics}
\label{sec:eval-metrics}
We evaluate an agent at three levels of granularity. At \textbf{component-level}, across all components, we report \textbf{tag prediction accuracy} ($Acc_{\text{tag}}$), the exact-match rate between the agent's predicted tag and the ground-truth tag, and \textbf{role description quality} ($Q_{\text{desc}}$), an LLM-judged score of the predicted role note against the ground-truth note. Specifically, the LLM-judge assesses the description quality on a 3-point scale (0 = incorrect, 1 = partially correct, 2 = correct). We use this scale to distinguish fully incorrect descriptions from partially correct ones that capture the main role but contain incorrect mechanistic sub-claims (an example is provided in Appendix \ref{app:qdesc-ex}). We then rescale the score to $[0, 1]$ as $Q_{\text{desc}}$.
At the \textbf{task-level}, for each task, we report \textbf{derived task accuracy} ($Acc_{\text{task}}$), a {binary} LLM-judged score of the agent's derived task description against the ground-truth task description. 
Finally, at the \textbf{process-level}, {we report \textbf{code execution success rate} ($S_{\text{exec}}$), the fraction of \tool{execute\_python} calls that run without error.}

{
LLM-judged metrics ($Q_{\text{desc}}$ and $Acc_{\text{task}}$) are scored independently by two LLM judges, GPT-5.4 and Gemini-3.1-Pro. We aggregate the two scores by taking the \textbf{lower} score instead of the mean. This choice provides a conservative estimate of explanation quality, which fits our setting because over-crediting an incorrect mechanistic claim is more harmful than under-crediting an incomplete one. The lower-score aggregation also reduces the impact of self-preference bias, where LLM judges can favor their own generations \citep{panickssery2024llmevaluatorsrecognizefavor}. A high score is retained only when both judges assign it.} 
\begin{figure*}[t!]
    \centering
    \includegraphics[width=0.95\linewidth]{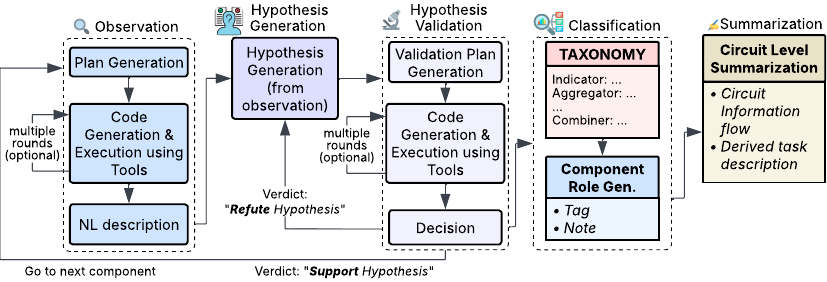}
    \caption{\textbf{\agent's pipeline}. \agent explains each localized component through an iterative \textit{observe} $\rightarrow$ \textit{hypothesize} $\rightarrow$ \textit{validate} loop. Refuted hypotheses are fed back as additional context to the next round. After processing all components, it assigns role tags and synthesizes a circuit-level summary. 
    }
    \label{fig:pipeline}
\end{figure*}

To validate the LLM-judged metrics, we collect human ratings on a subset of 10 tasks containing 17 components. For each component, two human judges independently evaluate the outputs of all four \agent backbones (Section~\ref{sec:results}), with agent identities hidden and randomized. This results in a total of 68 component-level annotations. The judges score $Q_{\text{desc}}$ and $Acc_{\text{task}}$ using the same rubrics as the LLM judges, and we observed a Cohen's $\kappa$ of $0.83$ for $Q_{\text{desc}}$ and $0.96$ for $Acc_{\text{task}}$, indicating almost perfect inter-annotator agreement \citep{landis1977observer}.  
As with the two LLM judges, we consider the lower score between the two human annotators as the ground-truth evaluation label, and report the LLM-human agreement. We observed substantial agreement for $Q_{\text{desc}}$ ($\kappa$=$0.76$) and almost perfect agreement for $Acc_{\text{task}}$ ($\kappa$=$0.8$), which confirms the validity of the LLM-judged metrics. We include details in Appendix~\ref{app:eval-details}.

\section{\agent}
\label{sec:pipeline}

In this section, we introduce \agent, our LM agent for circuit explanation.

\subsection{Overview}
\textbf{\agent} operates one component at a time. For each component in the localized circuit, it runs a three-stage analysis: \textbf{observe, hypothesize, and validate}. These three stages form an iterative loop. \agent generates a hypothesis from its observations, designs a controlled intervention to test it, and decides whether the evidence supports or refutes the claim. If refuted, the loop returns to hypothesis generation with the refuted claim as additional context. After all components are processed, \agent classifies each component, produces a component-level explanation, and derives the task description. Figure~\ref{fig:pipeline} illustrates the full pipeline.

\subsection{Observation}
The goal of \textbf{observation} is to gather descriptive evidence about a target component before hypothesizing its role. \agent first writes a structured observation plan specifying the goal of the observation, a step-by-step procedure, the required model tensors, and the expected pattern in the result. It then writes Python code implementing the plan, executes the code, and summarizes the results as a natural-language observation. We provide a helper library, \texttt{observation\_tools.py}, with primitives for inspecting attention patterns and activations. \agent may use these helpers, write custom code, or combine both.
\subsection{Hypothesis Generation}
\begin{table*}[!t]
    \centering
    \small
    \begin{tabular}{ll cccc}
    \toprule
    \textbf{Granularity} & \textbf{Metric} & \textbf{GPT-5.4} & \textbf{Claude-Sonnet-4.6} & \textbf{Gemini-3.1 Pro} & \textbf{Qwen-3-Coder-30B-A3B} \\
    \midrule
    
    \multirow{2}{*}{Component-level} 
      & $ Acc_{\text{tag}}$ & 0.74 & \textbf{0.79} & 0.76 & 0.67 \\
      % & $ Q_{\text{desc}}$ (0 -- 2) & 0.92 & 1.16 & \textbf{1.18} & 0.49\\
      & $Q_{\text{desc}}$ & 0.46 & 0.58 & \textbf{0.59} & 0.25 \\ % SCALED (divided by 2)
    \cmidrule(lr){1-6}
          
    Task-level
      & $ Acc_{\text{task}} $ & 0.63 & 0.75 & \textbf{0.83} & 0.25 \\
    \cmidrule(lr){1-6}
    % \multirow{2}{*}{Process-level}
    %   & $ \text{Val}_{\text{sound}}$ (0 -- 2)  & \textbf{1.91} & 1.32 & 1.49 & 0.91\\
    %   & $ S_{\text{exec}} $ (0 -- 1) & 0.52 & \textbf{0.93} & 0.80 & 0.62\\
    
    Process-level
      & $ S_{\text{exec}}$ & 0.52 & \textbf{0.93} & 0.80 & 0.62\\

    \bottomrule
    \end{tabular}
    \caption{Results of \agent with different LM backbones on \benchmark. 
    % $Acc_{\text{tag}}$, $S_{\text{exec}}$, and $Acc_{\text{task}}$ are reported on a 0 -- 1 scale, while $Q_{\text{desc}}$ and $\text{Val}_{\text{sound}}$ are 3-point scores on a 0 -- 2 scale. LLM-judged metrics are averaged over two judges, GPT-5.4 and Gemini-3.1-Pro. 
    Higher is better for all metrics. Bold indicates the best score in each row. 
    }
    \label{tab:headline}
\end{table*}

% Mean Numbers:
% \begin{tabular}{ll cccc}
% \toprule
% \textbf{Granularity} & \textbf{Metric (Scale)} & \textbf{GPT-5.4} & \textbf{Claude-Sonnet-4.6} & \textbf{Gemini-3.1 Pro} & \textbf{Qwen-3-Coder-30B-A3B} \\
% \midrule

% \multirow{2}{*}{Component-level} 
%   & $ Acc_{\text{tag}}$ (0 -- 1)& 0.74 & \textbf{0.79} & 0.76 & 0.67 \\
%   & $ Q_{\text{desc}}$ (0 -- 2) & 1.07  & \textbf{1.27} & 1.23 & 0.55\\
% \cmidrule(lr){1-6}
      
% Task-level
%   & $ Acc_{\text{task}} $ (0 -- 1) & 0.70 & 0.77 & \textbf{0.88} & 0.27 \\
% \cmidrule(lr){1-6}
% % \multirow{2}{*}{Process-level}
% %   & $ \text{Val}_{\text{sound}}$ (0 -- 2)  & \textbf{1.91} & 1.32 & 1.49 & 0.91\\
% %   & $ S_{\text{exec}} $ (0 -- 1) & 0.52 & \textbf{0.93} & 0.80 & 0.62\\

% Process-level
%   & $ S_{\text{exec}} $ (0 -- 1) & 0.52 & \textbf{0.93} & 0.80 & 0.62\\

% \bottomrule
% \end{tabular}

After observation, \agent proposes a hypothesis about the target component's role. The hypothesis is a short natural-language claim grounded in the observation, the task input-output examples, and any previously refuted hypotheses. At this stage, the role taxonomy is withheld, allowing \agent to reason freely about the component's behavior before committing to a fixed label. 
\subsection{Hypothesis Validation}
After a hypothesis is proposed, \agent tests it through controlled interventions on the target component. It first writes a structured validation plan specifying the prediction being tested, a step-by-step procedure, the activations or hooks to intervene on, and the result that would support or refute the hypothesis. It then writes Python code implementing the plan, executes the code, and issues a binary decision based on the results. We provide \texttt{validation\_tools.py}, a helper library with primitives for ablation, activation patching, and interchange interventions. Similar to the observation stage, \agent is free to use these primitives, write custom code, or combine both.
If the evidence supports the hypothesis, \agent moves on to the next component. If the evidence refutes it, the loop returns to hypothesis generation with the refuted claim added to the context, and \agent proposes a revised hypothesis informed by what has been ruled out.
\subsection{Classification}
{After all components have been processed}, \agent assigns each one a \textbf{\textit{tag}} from the taxonomy and writes a concise task-specific \textbf{\textit{note}} describing the role of each component based on the validated hypotheses. {At this stage, it} receives the taxonomy together with the final hypothesis for each component and selects the tag that best matches the component's role in the circuit. Separating classification from hypothesis generation allows \agent to reason about each component's behavior before committing to a fixed label. Introducing the taxonomy earlier could produce more taxonomy-aligned descriptions, but it would also constrain the agent's reasoning to the available tags, which we deliberately avoid. As we note in Section~\ref{sec:benchmark}, the taxonomy was introduced only for benchmarking; our workflow design enables \agent to work in practical applications where taxonomy does not exist.
\subsection{Summarization}
After classification, \agent synthesizes the component-level explanations into a circuit-level account of how the localized circuit implements the task. The summary contains two parts: First, a short description of how information flows between components, which serves as an intermediate step that externalizes \agent's findings; Second, a \textit{derived task description} inferred from the validated hypotheses and the task input-output examples. Only the derived task description is evaluated. It tests whether \agent can move beyond isolated component labels and recover the behavior implemented by the circuit as a whole, given that no task description was provided. 

\subsection{Implementation}
\agent is implemented as a graph-based state machine using LangGraph \citep{langgraph}. \textbf{Observation} and \textbf{Hypothesis Validation} stages share the same tool-calling procedure: \tool{list\_directory} and \tool{read\_file} for inspecting the helper libraries, built using TransformerLens \citep{nanda2022transformerlens}, and \tool{execute\_python} for running generated code. If the code execution fails, the model receives the error message and may revise its code. We allow up to five execution attempts per stage, after which the tool loop terminates and \agent must conclude with the evidence gathered so far. Generated code runs in a sandboxed subprocess against a pre-loaded LM. The hypothesis generation and validation loop is capped at three iterations per component. If the budget is exhausted without a supported hypothesis, \agent proceeds to the next component and retains its most recent hypothesis as a tentative explanation.

{We provide the reproducible prompts for \agent in Appendix~\ref{sec:appendix-prompts} and trace \agent's full trajectory for the running example in Appendix \ref{app:agent-walkthrough}}.

\section{Experiments}
\label{sec:results}

\subsection{Main Experimental Results}
We evaluate four frontier LLMs as agent backbones: GPT-5.4, Claude-Sonnet-4.6, Gemini-3.1-Pro, and Qwen-3-Coder-30B-A3B-Instruct.
Table~\ref{tab:headline} reports their results on \benchmark.

\paragraph{\agent provides meaningful circuit explanations, but no backbone dominates.}
Table~\ref{tab:headline} shows that \agent provides useful component- and task-level explanations, with different strengths across backbones. Claude-Sonnet-4.6 is strongest on component tagging and code execution, reaching $0.79$ $Acc_{\text{tag}}$ and $0.93$ $S_{\text{exec}}$. Gemini-3.1 Pro gives the best judged explanations, with the highest $Q_{\text{desc}}$ and $Acc_{\text{task}}$; its task accuracy is 8 points higher than the second-best backbone. GPT-5.4 remains competitive on tag prediction, but its low code execution success appears to limit its final explanation quality. Qwen-3-Coder trails the closed-weight models on the final explanation metrics. {Per-judge scores and repeated-run results on a fixed subset preserve these backbone-level trends (Appendices~\ref{app:judge-bias} and~\ref{app:run-stability}).}

\paragraph{Stronger LM backbones generate observation-grounded hypotheses.} 

We further analyze whether \agent's hypotheses follow from its own observations. On the 10-task, 17-component subset used for human validation, we manually rate each \textit{observation-hypothesis pair} for all four \agent backbones on a 0--2 grounding scale (0: hypotheses contradicting or ignoring observations; 1: hypotheses partially supported by observations; 2: fully supported). We include annotation details in Appendix~\ref{app:eval-details} and show examples in Table~\ref{tab:partial}.

All proprietary models reveal high consistency between observations and hypotheses (average score of $1.94$ with only one partially supported hypothesis and no ungrounded hypotheses). 
Qwen-3-Coder is lower, with a mean score of $1.41$ and only $41.2\%$ fully grounded hypotheses. It often starts from a valid but generic observation, but then over-specifies the hypothesis by adding unsupported task-specific mechanisms, such as particular per-neuron roles or positional rules. 

\paragraph{GPT-5.4 produces the soundest validation plans.}
{Given a grounded hypothesis, we ask whether \agent proposes an experiment that actually tests it. We manually score \emph{validation-plan soundness} on a 0-2 scale, ranging from no validation (0), indirect or incomplete validation (1), to full validation (2); example in Table \ref{tab:partial}.
The score judges only whether the proposed experiment would meaningfully support or refute the hypothesis, not whether the hypothesis itself is correct. Similar to Section~\ref{sec:eval-metrics}, we collect human ratings on the 10-task, 17-component subset and aggregate them as the lower of the two annotators' scores. GPT-5.4 is strongest with a score of $1.71$, followed by Gemini-3.1 Pro ($1.41$), Claude-Sonnet-4.6 ($1.24$), and Qwen-3-Coder ($0.71$). We provide scoring details and qualitative rubric examples in Appendix~\ref{app:eval-details}.
}
\newcommand{\partialmark}{\textcolor{orange}{\ding{115}}}
\newcommand{\hlpink}[1]{%
{\setlength{\fboxsep}{1pt}\colorbox{pink}{#1}}%
}
\newcommand{\hllime}[1]{%
{\setlength{\fboxsep}{1pt}\colorbox{lime}{#1}}%
}
\begin{table*}[!t]
    \centering
    \small
 \setlength{\tabcolsep}{5pt}
 \setlength{\extrarowheight}{2pt}
    \renewcommand{\arraystretch}{1.15}
    \begin{tabularx}{\textwidth}{
        >{\raggedright\arraybackslash}p{0.16\textwidth}
>{\raggedright\arraybackslash}p{0.12\textwidth}
>{\raggedright\arraybackslash}p{0.16\textwidth}
>{\raggedright\arraybackslash}X
    }
    \toprule
    \textbf{Task} & \textbf{Comp. (role)} & \textbf{Observation} & \textbf{Hypothesis loop} \\
    \midrule
    returns fraction of previous `x' tokens &
    \multirow{2}{=}{L0\_MLP (detect `x' tokens) }& 
    \multirow{2}{=}{\textbf{\textit{Obs:}}$\dots$ L2 Norm of MLP outputs vary between `x' and `non-x' tokens $\dots$ }
    & \textcolor{darkgreen}{\checkmark{}} \textit{\textbf{H1:}} L0\_MLP is a binary `is\_x' feature detector \textit{(\textbf{H1} is fully grounded in \textbf{Obs})} \\
    \textit{example}: (`c', `x', `a') $\rightarrow$ (0, 1/2, 1/3)& & & \textcolor{darkgreen}{\checkmark{}}\textbf{\textit{VP:}} Patch L0\_MLP activations among `x' and `non-x' in both directions. Outputs shift as if the token's \texttt{is\_x} value flipped. 
    \textit{(\textbf{VP} Tests all claims in \textbf{H1})}
    \\
    \midrule
    \multirow{2}{=}{Detects spam keywords. \newline
    \textit{example}: (`Hi', `offer', `free') $\rightarrow$ (`not spam', `spam', `spam')} & \multirow{2}{=}{L0\_MLP (detect each token from spam keywords \& emit per position signal)} & \multirow{2}{=}{\textbf{\textit{Obs:}}$\dots$ L0\_MLP has a high, stable activation norms across positions, dominated by small set of neurons $\dots$} & 
    \partialmark \textbf{\textit{H1:}} Detects \hllime{position-specific} spam patterns \hlpink{by aggregating features} from prev. tokens. (\textit{\textbf{H1} is partially-grounded in \textbf{Obs.}}) \\
    & & & \partialmark \textbf{\textit{VP: }}Patch L0\_MLP \hllime{on spam positions}, mean-ablate neuron 31, expect performance drops. (\textit{\textbf{VP} ignores the \textbf{aggregation} claim in \textbf{H1}})\\
    \midrule
    \multirow{2}{=}{Multiply each element by the sequence length
   \textit{example}: (2, 4, 6) $\rightarrow$ (6, 12, 18)} & 
    \multirow{2}{=}{L0\_MLP (computes per position seq. length from aggregation)} &
    \multirow{2}{=}{\textbf{\textit{Obs:}}$\dots$L0\_MLP has high activation norms, with position-dependent top neurons$\dots$} & 
    \textcolor{darkred}{\ding{55}{}}\textbf{\textit{H1:}} L0\_MLP applies a \hlpink{non-linear transformation} to each token.
    (\textit{\textbf{H1} not grounded in \textbf{Obs.}})\\
    & & &\textcolor{darkred}{\ding{55}{}}\textbf{\textit{VP: }} \hlpink{Mean ablate top 3 neurons, Test neuron} 
    \hlpink{84 for causal effect} (\textit{\textbf{VP} does not verify \textbf{any} claims in \textbf{H1}})
    \\
    \bottomrule
    \end{tabularx}
    \caption{
    Examples of hypothesis grounding and validation-plan soundness on benchmark tasks. Each row shows a task, an agent observation, the hypothesis (\textit{\textbf{H1}}), and the corresponding validation plan (\textit{\textbf{VP}}). (\textcolor{darkgreen}{\checkmark{}}) indicates grounded/sound, (\partialmark{}) indicates partial cases, and (\textcolor{darkred}{\ding{55}{}}) indicates ungrounded/unsound.
    \hlpink{Pink} marks the negative claims and \hllime{Green} marks the positive claims. 
    }
    \label{tab:partial}
\end{table*}

\paragraph{Reliable validation requires both sound plans and executable code.}
Sound validation plans are not sufficient unless they can be executed. GPT-5.4 has the strongest validation-plan ratings, but its low code execution success ($S_{\text{exec}}=0.52$) limits how often those plans yield usable evidence. Claude-Sonnet-4.6 shows the opposite pattern ($S_{\text{exec}}=0.93$), with reliable execution but weaker validation plans. Gemini-3.1 Pro is more balanced across the two dimensions, which helps explain its strong judged explanation scores.
\par To understand execution failures, we cluster the failed
\tool{execute\_python} calls into broad error categories. Figure~\ref{fig:code-error} shows that Python and tensor-manipulation bugs are common across backbones. Agents often make tensor shape mistakes, misuse helper or TransformerLens APIs, mishandle \verb*|<BOS>| offsets, or violate the tool protocol by omitting the required \texttt{result} variable. These patterns suggest that better execution scaffolding and more constrained helper APIs could improve \agent without changing the high-level reasoning loop.
\begin{figure}[h]
    \centering
    \includegraphics[width=0.95\linewidth]{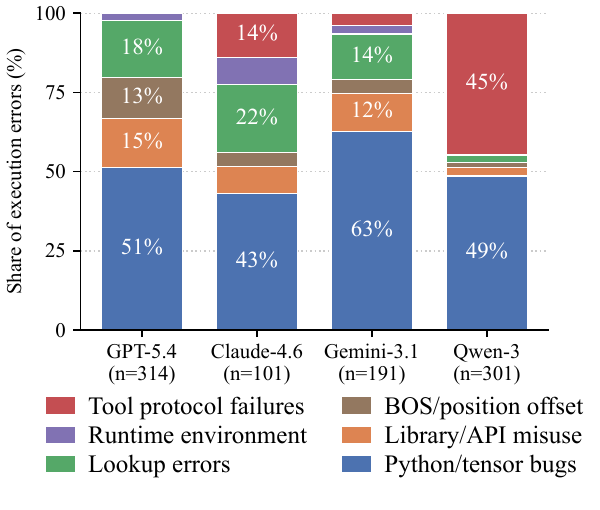}
    \caption{Distribution of \tool{execute\_python} errors by failure category and agent backbone.}
    \label{fig:code-error}
\end{figure}

\paragraph{Explanations improve when hypotheses converge.}
We examine the convergence rate of \agent backbones. This metric summarizes the downstream effect of the preceding failure modes: a grounded hypothesis must still be tested by a sound validation plan and executed successfully.
Figure~\ref{fig:hypothesis-convergence} shows that backbones with fewer unresolved components tend to achieve stronger
final explanations: Claude-Sonnet-4.6 converges most reliably and attains the best tag accuracy, while Qwen-3-Coder leaves many components unresolved and performs worst. Despite producing the soundest validation plans, GPT-5.4 has the lowest convergence rate among the proprietary models, as many of its validation attempts fail at execution. Thus, a sound plan improves final explanations only when the agent can execute it and turn the result into usable evidence. This suggests that final explanation quality depends on completing the full \textit{observe, hypothesize, validate} loop.
We provide token usage and estimated API cost for running \agent with each backbone in Appendix~\ref{app:api-cost}.
\begin{figure}[h]
\centering
\includegraphics[width=\columnwidth]{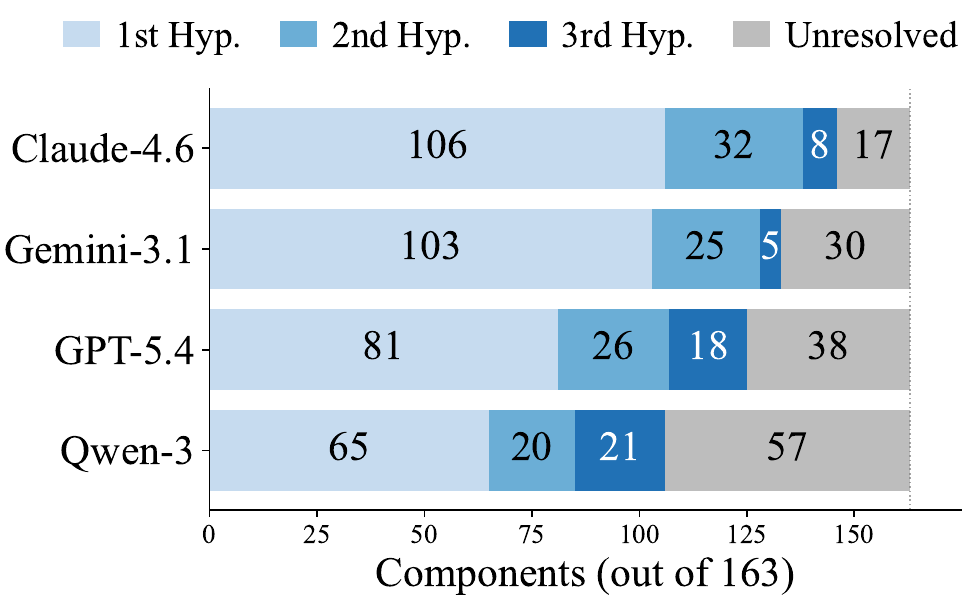}
\caption{
Hypothesis convergence per backbone. Each bar shows the number of components supported on the 1st, 2nd, or 3rd hypothesis-generation iteration, or left unresolved after the three-iteration budget.
}
\label{fig:hypothesis-convergence}
\end{figure}

{
\subsection{Workflow Ablations}
We test the contribution of individual \agent stages through two ablations on the full benchmark. \textsc{No-observation} removes the dedicated observation stage but retains hypothesis generation and validation. \textsc{No-validation} removes the validation loop and passes the initial hypothesis directly to classification and summarization. Table~\ref{tab:workflow-ablations} shows that the full pipeline achieves the highest ${Acc_\text{task}}$ for every backbone under both ablations. Removing either stage decreases ${Acc_\text{task}}$ by 1--18 percentage points. The full pipeline also achieves higher component-level scores in 14 of the 16 comparisons. These results indicate that both stages contribute to the final explanations beyond the capability of the underlying backbone. We further note that \textsc{No-validation} is a \textit{diagnostic ablation only}. Even when an initial hypothesis produces a correct final description, causal validation is necessary before accepting it as a mechanistic explanation.}
\begin{table}[h]
\centering
\small
\setlength{\tabcolsep}{3pt}
\begin{tabular}{llcccc}
\toprule
Backbone & System & $Acc_{\text{tag}}$ & $Q_{\text{desc}}$
& $Acc_{\text{task}}$ & $S_{\text{exec}}$ \\
\midrule
GPT-5.4 & Full \agent & 0.74 & 0.46 & 0.63 & 0.52 \\
& No-obs & 0.66 & 0.47 & 0.59 & 0.52 \\
& No-val & 0.70 & 0.45 & 0.60 & 0.52 \\
\midrule
Claude & Full \agent & 0.79 & 0.58 & 0.75 & 0.93 \\
& No-obs & 0.72 & 0.50 & 0.57 & 0.94 \\
& No-val & 0.73 & 0.51 & 0.65 & 0.94 \\
\midrule
Gemini & Full \agent & 0.76 & 0.59 & 0.83 & 0.80 \\
& No-obs & 0.73 & 0.52 & 0.73 & 0.61 \\
& No-val & 0.69 & 0.53 & 0.69 & 0.80 \\
\midrule
Qwen & Full \agent & 0.67 & 0.25 & 0.25 & 0.62 \\
& No-obs & 0.65 & 0.21 & 0.23 & 0.46 \\
& No-val & 0.68 & 0.24 & 0.24 & 0.85 \\
\bottomrule
\end{tabular}
\caption{Results of \agent stage ablations on the full \benchmark.}
\label{tab:workflow-ablations}
\end{table}

{We next compare \agent with Claude Code to test whether our structured pipeline offers benefits beyond unrestricted tool-using agent behavior. We run Claude Code with Claude-Sonnet-4.6 on a 31-task subset containing 55 components, providing the same context as how we evaluated \agent, except the imposed workflow. 
% Claude Code receives the same task descriptions, circuit files, input-output examples, tools, and output requirements as \agent, but without the imposed workflow. 
Table~\ref{tab:claude-code-baseline} shows that \agent improves $Acc_{\text{tag}}$, $Q_{\text{desc}}$, and $S_{\text{exec}}$ while matching $Acc_{\text{task}}$. Manual inspection shows that Claude Code performs a causal intervention in only 15 of 31 cases; 12 rely only on observational evidence, and 4 infer the task from examples without loading the model. Thus, Claude Code often produces explanations without testing whether the inferred roles are causally supported, a key requirement for rigorous mechanistic interpretation.}

\begin{table}[h]
\centering
\small
\setlength{\tabcolsep}{4pt}
\begin{tabular}{lcccc}
\toprule
System & $Acc_{\text{tag}}$ & $Q_{\text{desc}}$
& $Acc_{\text{task}}$ & $S_{\text{exec}}$ \\
\midrule
\agent (Claude) & 0.811 & 0.639 & 0.767 & 0.94 \\
Claude Code     & 0.770 & 0.603 & 0.767 & 0.69 \\
\bottomrule
\end{tabular}
\caption{Comparison between \agent and Claude Code on a 31-task/55-component subset.}
\label{tab:claude-code-baseline}
\end{table}

\section{A Case Study on Realistic LM}
\benchmark provides controlled ground truth by relying on semi-synthetic transformers, but these models do not capture the full setting of a naturally trained, large-scale autoregressive LM. To test whether \agent's behavior carries over to this setting, we conduct a case study on the All-for-One (AF1) circuit identified by \citet{mamidanna-etal-2025-one} for the three-operand task $A+B+C$ in Llama-3-8B \citep{llama3modelcard}.\footnote{The AF1 paper was published after the reported training-data cutoffs of the proprietary backbones we use, reducing the likelihood of data leakage.} Compared to \benchmark, this setting introduces additional challenges: (i) a larger localized circuit with more components, (ii) redundant routes between attention heads leading to backup heads, and (iii) components that appear important under logit lens probes but are causally weak under intervention.
{
The localized circuit contains 10 components, including operand-transfer attention heads, late layer MLPs, and logit-lens-positive attention heads. We manually construct
component-level reference roles using targeted interventions, retaining both causal and redundant components to test whether \agent can distinguish mechanistic evidence from suggestive but non-causal signals.
We provide setup and reference-annotation details in Appendix~\ref{sec:appendix-realckt}.}

We run \agent with three LM backbones: GPT-5.4, Claude-Sonnet-4.6, and Gemini-3.1-Pro. {We omit Qwen-3-Coder as it substantially underperforms the closed-weight models on component descriptions and task inference in the controlled benchmark.} 
Three human annotators independently rate the natural-language role note produced by each agent and we report a majority vote.

Table~\ref{tab:af1-results} shows the performance across these models. Claude-Sonnet-4.6 and GPT-5.4 generally recover the transfer-head structure and distinguish causally redundant late components from necessary ones {(Claude-Sonnet-4.6-\agent iteration example in Appendix Table~\ref{tab:agent-exploration})}. The main failure mode is over-interpreting answer-correlated evidence: Gemini-3.1-Pro often treats positional or logit-lens signals as causal, leading to incorrect role descriptions.
{More broadly, this case study suggests a practical use for \agent: applying it to already studied realistic circuits could test whether agentic explainers can reproduce known human findings and even surface certain overlooked mechanisms.
\begin{table}[h]
    \centering
    \small
    \begin{tabular}{lccc}
    \toprule
    Agent & Correct & Partial & Wrong \\
    \midrule
    GPT-5.4 & 6 & 3 & 1 \\
    Claude-Sonnet-4.6 & 8 & 2 & 0 \\
    Gemini-3.1-Pro & 1 & 2 & 7 \\
    \bottomrule
    \end{tabular}
    \caption{Human-rated role-description quality on the 10-component AF1 circuit of Llama-3-8B.}
    \label{tab:af1-results}
\end{table}
}

\section{Conclusion} \label{sec:conclusion-future-work}

We study whether LM agents can explain localized circuits in transformers. To this end, we introduce a controlled benchmark and a new agentic circuit explanation framework.
Our results show that LM agents can produce useful circuit explanations, but the problem is not solved. Stronger backbones usually generate grounded hypotheses. The harder step is validating them through sound causal tests and reliable code execution. This validation loop is where failures occur, especially through incomplete validation plans and code execution errors. 

{Our findings suggest that the immediate practical value of agentic circuit explanation lies in assisting human researchers. Given a localized circuit, an agent can propose component roles, design interventions to test them, and investigate potentially overlooked mechanisms. However, plausible explanations should not be accepted without successful causal validation, and incomplete or failed experiments should remain visible for human review.}

\section*{Limitations and Future Work}
This work evaluates circuit explanation in a post-localization setting. \agent is given the localized circuit and asked to explain its components. Thus, our results measure the explanation stage rather than end-to-end circuit discovery.

\benchmark uses semi-synthetic circuits with recoverable ground truth. This enables systematic evaluation, but the circuits are smaller, more structured, and more algorithmic than many mechanisms in naturally trained LMs. Our real-model case study provides an initial test beyond this setting. Future work should evaluate larger and more naturally occurring circuits while accounting for potential data leakage, which can invalidate benchmarking results. 

The results reflect one agent design, prompting setup, and helper library for code execution. Richer helper libraries and a more customized harness, such as more constrained execution interfaces and alternative interaction designs, may reduce code-level failures and make causal interventions easier for future agents.

\section*{Acknowledgments}
We appreciate the sponsorship from Foresight Institute. This project was also supported by resources provided by the Office of Research Computing at George Mason University (URL: https://orc.gmu.edu) and funded in part by grants from the National Science Foundation (Award Number 2018631).

\bibliography{anthology-1, anthology-2, custom, yao}

\begin{thebibliography}{37}
\providecommand{\natexlab}[1]{#1}

\bibitem[{AI@Meta(2024)}]{llama3modelcard}
AI@Meta. 2024.
\newblock \href {https://github.com/meta-llama/llama3/blob/main/MODEL_CARD.md} {Llama 3 model card}.

\bibitem[{{Anthropic}(2026)}]{anthropic2026claudesonnet46}
{Anthropic}. 2026.
\newblock Claude sonnet 4.6 system card.
\newblock \url{https://www-cdn.anthropic.com/78073f739564e986ff3e28522761a7a0b4484f84.pdf}.
\newblock Accessed: 2026-05-22.

\bibitem[{Bai et~al.(2026)Bai, Baumgartner, Sun, Holtzman, and Tan}]{bai2026storyscienceexecutiongroundedevaluation}
Xiaoyan Bai, Alexander Baumgartner, Haojia Sun, Ari Holtzman, and Chenhao Tan. 2026.
\newblock \href {https://arxiv.org/abs/2602.18458} {The story is not the science: Execution-grounded evaluation of mechanistic interpretability research}.
\newblock \emph{Preprint}, arXiv:2602.18458.

\bibitem[{Bereska and Gavves(2024)}]{bereska2024mechanistic}
Leonard Bereska and Efstratios Gavves. 2024.
\newblock Mechanistic interpretability for ai safety--a review.
\newblock \emph{arXiv preprint arXiv:2404.14082}.

\bibitem[{Bissell et~al.(2025)Bissell, Byun, and Balsam}]{bissell2025researchagent}
Mark Bissell, Michael Byun, and Daniel Balsam. 2025.
\newblock \href {https://www.goodfire.com/blog/you-and-your-research-agent} {You and your research agent: Lessons from using agents for interpretability research}.
\newblock \emph{Goodfire Research}.

\bibitem[{Chen et~al.(2025)Chen, Chen, Ning, Zhang, Wang, Yu, Li, Liao, Wei, Lu, Dey, Xue, Baker, Burns, Adu-Ampratwum, Huang, Ning, Gao, Su, and Sun}]{chen2025scienceagentbench}
Ziru Chen, Shijie Chen, Yuting Ning, Qianheng Zhang, Boshi Wang, Botao Yu, Yifei Li, Zeyi Liao, Chen Wei, Zitong Lu, Vishal Dey, Mingyi Xue, Frazier~N. Baker, Benjamin Burns, Daniel Adu-Ampratwum, Xuhui Huang, Xia Ning, Song Gao, Yu~Su, and Huan Sun. 2025.
\newblock \href {https://openreview.net/forum?id=6z4YKr0GK6} {Scienceagentbench: Toward rigorous assessment of language agents for data-driven scientific discovery}.
\newblock In \emph{The Thirteenth International Conference on Learning Representations}.

\bibitem[{Conmy et~al.(2023)Conmy, Mavor-Parker, Lynch, Heimersheim, and Garriga-Alonso}]{conmy2023towards}
Arthur Conmy, Augustine Mavor-Parker, Aengus Lynch, Stefan Heimersheim, and Adri{\`a} Garriga-Alonso. 2023.
\newblock Towards automated circuit discovery for mechanistic interpretability.
\newblock \emph{Advances in Neural Information Processing Systems}, 36:16318--16352.

\bibitem[{Ferrando et~al.(2024)Ferrando, Sarti, Bisazza, and Costa-jussà}]{ferrando2024primerinnerworkingstransformerbased}
Javier Ferrando, Gabriele Sarti, Arianna Bisazza, and Marta~R. Costa-jussà. 2024.
\newblock \href {https://arxiv.org/abs/2405.00208} {A primer on the inner workings of transformer-based language models}.
\newblock \emph{Preprint}, arXiv:2405.00208.

\bibitem[{Geiger et~al.(2022)Geiger, Wu, Lu, Rozner, Kreiss, Icard, Goodman, and Potts}]{pmlr-v162-geiger22a}
Atticus Geiger, Zhengxuan Wu, Hanson Lu, Josh Rozner, Elisa Kreiss, Thomas Icard, Noah Goodman, and Christopher Potts. 2022.
\newblock \href {https://proceedings.mlr.press/v162/geiger22a.html} {Inducing causal structure for interpretable neural networks}.
\newblock In \emph{Proceedings of the 39th International Conference on Machine Learning}, volume 162 of \emph{Proceedings of Machine Learning Research}, pages 7324--7338. PMLR.

\bibitem[{{Google DeepMind}(2026)}]{googledeepmind2026gemini31pro}
{Google DeepMind}. 2026.
\newblock Gemini 3.1 pro model card.
\newblock \url{https://storage.googleapis.com/deepmind-media/Model-Cards/Gemini-3-1-Pro-Model-Card.pdf}.
\newblock Accessed: 2026-05-22.

\bibitem[{Gupta et~al.(2025)Gupta, Arcuschin, Kwa, and Garriga-Alonso}]{gupta2025interpbench}
Rohan Gupta, Iván Arcuschin, Thomas Kwa, and Adrià Garriga-Alonso. 2025.
\newblock \href {https://arxiv.org/abs/2407.14494} {Interpbench: Semi-synthetic transformers for evaluating mechanistic interpretability techniques}.
\newblock \emph{Preprint}, arXiv:2407.14494.

\bibitem[{Haklay et~al.(2026)Haklay, Prakash, Pandey, Torralba, Mueller, Andreas, Shaham, and Belinkov}]{haklay2026pitfallsevaluatinginterpretabilityagents}
Tal Haklay, Nikhil Prakash, Sana Pandey, Antonio Torralba, Aaron Mueller, Jacob Andreas, Tamar~Rott Shaham, and Yonatan Belinkov. 2026.
\newblock \href {https://arxiv.org/abs/2603.20101} {Pitfalls in evaluating interpretability agents}.
\newblock \emph{Preprint}, arXiv:2603.20101.

\bibitem[{Hanna et~al.(2023)Hanna, Liu, and Variengien}]{hanna2023doesgpt2computegreaterthan}
Michael Hanna, Ollie Liu, and Alexandre Variengien. 2023.
\newblock \href {https://arxiv.org/abs/2305.00586} {How does gpt-2 compute greater-than?: Interpreting mathematical abilities in a pre-trained language model}.
\newblock \emph{Preprint}, arXiv:2305.00586.

\bibitem[{Hanna et~al.(2024)Hanna, Pezzelle, and Belinkov}]{hanna2024faithfaithfulnessgoingcircuit}
Michael Hanna, Sandro Pezzelle, and Yonatan Belinkov. 2024.
\newblock \href {https://arxiv.org/abs/2403.17806} {Have faith in faithfulness: Going beyond circuit overlap when finding model mechanisms}.
\newblock \emph{Preprint}, arXiv:2403.17806.

\bibitem[{Kantamneni and Tegmark(2025)}]{kantamneni2025languagemodelsusetrigonometry}
Subhash Kantamneni and Max Tegmark. 2025.
\newblock \href {https://arxiv.org/abs/2502.00873} {Language models use trigonometry to do addition}.
\newblock \emph{Preprint}, arXiv:2502.00873.

\bibitem[{Landis and Koch(1977)}]{landis1977observer}
J.~Richard Landis and Gary~G. Koch. 1977.
\newblock The measurement of observer agreement for categorical data.
\newblock \emph{Biometrics}, 33(1):159--174.

\bibitem[{{LangChain AI}(2024)}]{langgraph}
{LangChain AI}. 2024.
\newblock Langgraph.
\newblock \url{https://github.com/langchain-ai/langgraph}.

\bibitem[{Lindner et~al.(2023)Lindner, Kramár, Farquhar, Rahtz, McGrath, and Mikulik}]{lindner2023tracrcompiledtransformerslaboratory}
David Lindner, János Kramár, Sebastian Farquhar, Matthew Rahtz, Thomas McGrath, and Vladimir Mikulik. 2023.
\newblock \href {https://arxiv.org/abs/2301.05062} {Tracr: Compiled transformers as a laboratory for interpretability}.
\newblock \emph{Preprint}, arXiv:2301.05062.

\bibitem[{Liu et~al.(2026)Liu, Miao, Zhao, Liu, and Du}]{liu2026neuronscopemultiagentframeworkexplaining}
Weiqi Liu, Yongliang Miao, Haiyan Zhao, Yanguang Liu, and Mengnan Du. 2026.
\newblock \href {https://arxiv.org/abs/2601.03671} {Neuronscope: A multi-agent framework for explaining polysemantic neurons in language models}.
\newblock \emph{Preprint}, arXiv:2601.03671.

\bibitem[{Lu et~al.(2024)Lu, Lu, Lange, Foerster, Clune, and Ha}]{lu2024ai}
Chris Lu, Cong Lu, Robert~Tjarko Lange, Jakob Foerster, Jeff Clune, and David Ha. 2024.
\newblock The ai scientist: Towards fully automated open-ended scientific discovery.
\newblock \emph{arXiv preprint arXiv:2408.06292}.

\bibitem[{Mamidanna et~al.(2025)Mamidanna, Rai, Yao, and Zhou}]{mamidanna-etal-2025-one}
Siddarth Mamidanna, Daking Rai, Ziyu Yao, and Yilun Zhou. 2025.
\newblock \href {https://doi.org/10.18653/v1/2025.emnlp-main.1565} {All for one: {LLM}s solve mental math at the last token with information transferred from other tokens}.
\newblock In \emph{Proceedings of the 2025 Conference on Empirical Methods in Natural Language Processing}, pages 30747--30760, Suzhou, China. Association for Computational Linguistics.

\bibitem[{Marin-Llobet and Ferrando(2026)}]{marinllobet2026automatedinterpretabilityfeaturediscovery}
Arnau Marin-Llobet and Javier Ferrando. 2026.
\newblock \href {https://arxiv.org/abs/2605.01555} {Automated interpretability and feature discovery in language models with agents}.
\newblock \emph{Preprint}, arXiv:2605.01555.

\bibitem[{Mueller et~al.(2025)Mueller, Geiger, Wiegreffe, Arad, Arcuschin, Belfki, Chan, Fiotto-Kaufman, Haklay, Hanna, Huang, Gupta, Nikankin, Orgad, Prakash, Reusch, Sankaranarayanan, Shao, Stolfo, Tutek, Zur, Bau, and Belinkov}]{pmlr-v267-mueller25a}
Aaron Mueller, Atticus Geiger, Sarah Wiegreffe, Dana Arad, Iv\'{a}n Arcuschin, Adam Belfki, Yik~Siu Chan, Jaden~Fried Fiotto-Kaufman, Tal Haklay, Michael Hanna, Jing Huang, Rohan Gupta, Yaniv Nikankin, Hadas Orgad, Nikhil Prakash, Anja Reusch, Aruna Sankaranarayanan, Shun Shao, Alessandro Stolfo, and 4 others. 2025.
\newblock \href {https://proceedings.mlr.press/v267/mueller25a.html} {{MIB}: A mechanistic interpretability benchmark}.
\newblock In \emph{Proceedings of the 42nd International Conference on Machine Learning}, volume 267 of \emph{Proceedings of Machine Learning Research}, pages 45069--45108. PMLR.

\bibitem[{Nanda and Bloom(2022)}]{nanda2022transformerlens}
Neel Nanda and Joseph Bloom. 2022.
\newblock Transformerlens.
\newblock \url{https://github.com/TransformerLensOrg/TransformerLens}.

\bibitem[{{OpenAI}(2026)}]{openai2026gpt54}
{OpenAI}. 2026.
\newblock Introducing gpt-5.4.
\newblock \url{https://openai.com/index/introducing-gpt-5-4/}.
\newblock Accessed: 2026-05-22.

\bibitem[{Panickssery et~al.(2024)Panickssery, Bowman, and Feng}]{panickssery2024llmevaluatorsrecognizefavor}
Arjun Panickssery, Samuel~R. Bowman, and Shi Feng. 2024.
\newblock \href {https://arxiv.org/abs/2404.13076} {Llm evaluators recognize and favor their own generations}.
\newblock \emph{Preprint}, arXiv:2404.13076.

\bibitem[{{Paulo} et~al.(2024){Paulo}, {Mallen}, {Juang}, and {Belrose}}]{2024arXiv241013928P}
Gon{\c{c}}alo {Paulo}, Alex {Mallen}, Caden {Juang}, and Nora {Belrose}. 2024.
\newblock \href {https://doi.org/10.48550/arXiv.2410.13928} {{Automatically Interpreting Millions of Features in Large Language Models}}.
\newblock \emph{arXiv e-prints}, arXiv:2410.13928.

\bibitem[{Qwen(2025)}]{qwen3technicalreport}
Qwen. 2025.
\newblock \href {https://arxiv.org/abs/2505.09388} {Qwen3 technical report}.
\newblock \emph{Preprint}, arXiv:2505.09388.

\bibitem[{Rai et~al.(2024)Rai, Zhou, Feng, Saparov, and Yao}]{rai2024practical}
Daking Rai, Yilun Zhou, Shi Feng, Abulhair Saparov, and Ziyu Yao. 2024.
\newblock A practical review of mechanistic interpretability for transformer-based language models.
\newblock \emph{arXiv preprint arXiv:2407.02646}.

\bibitem[{{Schwettmann} et~al.(2023){Schwettmann}, {Rott Shaham}, {Materzynska}, {Chowdhury}, {Li}, {Andreas}, {Bau}, and {Torralba}}]{2023arXiv230903886S}
Sarah {Schwettmann}, Tamar {Rott Shaham}, Joanna {Materzynska}, Neil {Chowdhury}, Shuang {Li}, Jacob {Andreas}, David {Bau}, and Antonio {Torralba}. 2023.
\newblock \href {https://doi.org/10.48550/arXiv.2309.03886} {{FIND: A Function Description Benchmark for Evaluating Interpretability Methods}}.
\newblock \emph{arXiv e-prints}, arXiv:2309.03886.

\bibitem[{Shaham et~al.(2024)Shaham, Schwettmann, Wang, Rajaram, Hernandez, Andreas, and Torralba}]{pmlr-v235-shaham24a}
Tamar~Rott Shaham, Sarah Schwettmann, Franklin Wang, Achyuta Rajaram, Evan Hernandez, Jacob Andreas, and Antonio Torralba. 2024.
\newblock \href {https://proceedings.mlr.press/v235/shaham24a.html} {A multimodal automated interpretability agent}.
\newblock In \emph{Proceedings of the 41st International Conference on Machine Learning}, volume 235 of \emph{Proceedings of Machine Learning Research}, pages 44293--44321. PMLR.

\bibitem[{Singh et~al.(2026)Singh, Tan, Xu, Gero, Yang, Galley, and Gao}]{singh2026agenticimodels}
Chandan Singh, Yan~Shuo Tan, Weijia Xu, Zelalem Gero, Weiwei Yang, Michel Galley, and Jianfeng Gao. 2026.
\newblock \href {https://arxiv.org/abs/2605.03808} {Agentic-imodels: Evolving agentic interpretability tools via autoresearch}.
\newblock \emph{Preprint}, arXiv:2605.03808.

\bibitem[{Syed et~al.(2024)Syed, Rager, and Conmy}]{syed2024attribution}
Aaquib Syed, Can Rager, and Arthur Conmy. 2024.
\newblock Attribution patching outperforms automated circuit discovery.
\newblock In \emph{Proceedings of the 7th BlackboxNLP Workshop: Analyzing and Interpreting Neural Networks for NLP}, pages 407--416.

\bibitem[{Thurnherr and Scheurer(2024)}]{thurnherr2024tracrbench}
Hannes Thurnherr and Jérémy Scheurer. 2024.
\newblock \href {https://arxiv.org/abs/2409.13714} {Tracrbench: Generating interpretability testbeds with large language models}.
\newblock \emph{Preprint}, arXiv:2409.13714.

\bibitem[{Wang et~al.(2023)Wang, Variengien, Conmy, Shlegeris, and Steinhardt}]{wang2023interpretability}
Kevin~Ro Wang, Alexandre Variengien, Arthur Conmy, Buck Shlegeris, and Jacob Steinhardt. 2023.
\newblock \href {https://openreview.net/forum?id=NpsVSN6o4ul} {Interpretability in the wild: a circuit for indirect object identification in {GPT}-2 small}.
\newblock In \emph{The Eleventh International Conference on Learning Representations}.

\bibitem[{Weiss et~al.(2021)Weiss, Goldberg, and Yahav}]{weiss2021thinking}
Gail Weiss, Yoav Goldberg, and Eran Yahav. 2021.
\newblock \href {https://arxiv.org/abs/2106.06981} {Thinking like transformers}.
\newblock \emph{Preprint}, arXiv:2106.06981.

\bibitem[{Yamada et~al.(2025)Yamada, Lange, Lu, Hu, Lu, Foerster, Clune, and Ha}]{yamada2025ai}
Yutaro Yamada, Robert~Tjarko Lange, Cong Lu, Shengran Hu, Chris Lu, Jakob Foerster, Jeff Clune, and David Ha. 2025.
\newblock The ai scientist-v2: Workshop-level automated scientific discovery via agentic tree search.
\newblock \emph{arXiv preprint arXiv:2504.08066}.

\end{thebibliography}

\newpage
\appendix
\section{Prompt Templates}
\label{sec:appendix-prompts}
In this section, we include the templates used to prompt \agent. 
\subsection{System Prompt}
\label{app:system-prompt}
\begin{tcolorbox}[promptbox, title=System Prompt]
You are a research agent specializing in mechanistic interpretability of transformer language models. Your goal is to explain how a localized circuit, and in particular each target component, contributes to the model's behavior on a given sequence-processing task.\\
\textbf{Context:}
\begin{enumerate} [itemsep=2pt, parsep=0pt]
\item \textbf{Task:} The model performs a sequence-processing task. See the Task Information below for input-output examples.
\item \textbf{Target model:} A decoder-only transformer with \texttt{\{NUM\_LAYERS\}} layers, \texttt{\{NUM\_HEADS\}} attention heads per layer, and an MLP sublayer in each block.
\item \textbf{Localized components:} \texttt{\{ALL\_COMPONENTS\}}
\item \textbf{Target component:} \texttt{\{COMPONENT\_IDENTIFIER\}}
\item \textbf{Component Localization:} The listed components have been identified as causally relevant to the task. Your goal is to explain the functional role of the target component within this circuit.
\item \textbf{Previously validated components:} \texttt{\{VALIDATED\}}.
\item \textbf{Iterative Workflow}:
\begin{enumerate}
    \item \textbf{Observation:} Gather descriptive evidence about the behavior and activation patterns of \texttt{{COMPONENT\_IDENTIFIER}} using the available analysis tools.
   \item \textbf{Hypothesis Generation:} Propose a hypothesis about the functional role of \texttt{{COMPONENT\_IDENTIFIER}}, grounded in your observations.
   \item \textbf{Hypothesis Validation:} Design and execute causal or behavioral tests to support or refute the hypothesis.
   \item \textbf{Classification:} After all components are validated, assign each a functional role tag from a fixed taxonomy and a concise functional role note, grounded in its validated hypothesis.
   \item \textbf{Summarization:} Once all components are classified, synthesize how the information flows through the localized circuit. Write a description of the task being performed by the model.
\end{enumerate}
\end{enumerate}

\textbf{Guidelines.}
Do not assume the role of any component in advance. Prefer precise, falsifiable hypotheses. Ground claims in mechanistic evidence from activations, interventions, and model behavior.
\end{tcolorbox}
\newpage
\subsection{Observation Stage}
\label{app:observation-prompt}
\begin{tcolorbox}[promptbox, title=Observation Plan Prompt]
You are currently in the \textbf{Observation} stage, \textbf{Plan Generation} step. Your goal is to gather descriptive evidence about \texttt{{COMPONENT\_IDENTIFIER}} that will guide hypothesis generation in the next stage.

Generate a structured plan specifying what to observe about \texttt{\{COMPONENT\_IDENTIFIER\}}. Follow these instructions:
\begin{enumerate}[itemsep=2pt, parsep=0pt]
\item The plan must be purely descriptive and observational. Do not include ablations or causal interventions.
\item Be concrete and implementable. After this step, Python code will be generated directly from this plan.
\item Keep each field concise, using 1-3 sentences.
\end{enumerate}
Analysis tools are available in the execution environment and will be discovered during code execution.
Respond with a JSON object containing exactly the following fields, and do not include any text outside the JSON:
\begin{itemize}[itemsep=2pt, parsep=0pt]
\item \texttt{stage}: \texttt{"OBSERVATION PLAN for \{COMPONENT\_IDENTIFIER\}"}
\item \texttt{goal}: what aspect of the component should be characterized, and why it is relevant.
\item \texttt{procedure}: a short list of concrete observation steps.
\item \texttt{inputs}: the model tensors, hooks, or cached activations required.
\item \texttt{expected\_result}: the expected pattern or value and how it should be interpreted.
\end{itemize}
\end{tcolorbox}
% \newpage
\begin{tcolorbox}[promptbox, title=Observation Code and Execution Prompt]
You are currently in the \textbf{Observation} stage, \textbf{Code Generation and Execution} step.

\textbf{Observation plan:} \texttt{{OBSERVATION\_PLAN}}

You have access to three tools:
\begin{itemize}[itemsep=2pt, parsep=0pt]
\item \tool{list\_directory(path)}: list files in a directory on the analysis server.
\item \tool{read\_file(path)}: read the full contents of a file on the analysis server.
\item \tool{execute\_python(code)}: execute Python code in the pre-configured execution environment.
\end{itemize}

The execution environment has the following objects in scope:
\begin{itemize}[itemsep=2pt, parsep=0pt]
\item \texttt{model}: a \texttt{HookedTransformer}. 
\item \texttt{tokenize(sequences)}: converts a list of token lists into a \texttt{[batch, seq\_len]} tensor. Each sequence starts with \texttt{BOS} and uses tokens from the task vocabulary.
\item \texttt{decode(logits)}: converts model output logits to task-semantic values in the same format as the task examples. The BOS position is excluded by default.
\item Standard imports
\end{itemize}

Recommended steps:
\begin{enumerate}[itemsep=2pt, parsep=0pt]
\item Call \tool{list\_directory()} to discover available helper files.
\item Read the file most relevant to the observation plan to learn the available function signatures. 
\item Generate executable Python code for the observation and call \tool{execute\_python}. The code must implement the observation plan, assign all outputs to a dictionary named \texttt{result}.
\end{enumerate}

Each \tool{execute\_python} call runs in a fresh subprocess. Variables, imports, and state do not persist across calls. Every call must be self-contained. Do not perform causal interventions. You have up to \texttt{\{MAX\_EXECUTE\_ROUNDS\}} execution attempts. If code returns an error, read the traceback, identify the root cause, and revise the next attempt.

When sufficient evidence has been gathered to characterize \texttt{\{COMPONENT\_IDENTIFIER\}}, respond with a JSON object containing exactly one field:
\begin{itemize}[itemsep=2pt, parsep=0pt]
\item \texttt{observation}: a concise natural-language description of what was observed about \texttt{\{COMPONENT\_IDENTIFIER\}}.
\end{itemize}
\end{tcolorbox}
% \newpage
\subsection{Hypothesis Generation Stage}
\label{app:hypothesis-generation-prompt}
\begin{tcolorbox}[promptbox, title=Hypothesis Generation Prompt]
You are currently in the \textbf{Hypothesis Generation} stage. Based on the observation collected in the previous stage (if available) and the task context provided in the system prompt, your goal is to propose a precise, falsifiable hypothesis about the functional role of \texttt{\{COMPONENT\_IDENTIFIER\}} in the circuit.

\textbf{Prior observation:} \texttt{\{OBSERVATION\}}

\textbf{Previously refuted hypotheses (if any):} \texttt{\{REFUTED\}}

\textbf{Instructions:}
\begin{enumerate}[itemsep=2pt, parsep=0pt]
\item Your hypothesis must be grounded in the prior observation. Do not contradict observed evidence.
\item If previous hypotheses were refuted, explicitly address why they failed and constrain the new proposal accordingly.
\item A good hypothesis specifies: (a) what the component computes, (b) which input positions or tokens it operates on, and (c) what it contributes downstream.
\item Use language that reflects a testable claim, such as ``we hypothesize that \dots''
\end{enumerate}

Respond with a JSON object containing exactly the following fields, and do not include any text outside the JSON:
\begin{itemize}[itemsep=2pt, parsep=0pt]
\item \texttt{hypothesis}: the proposed hypothesis about the functional role of \texttt{\{COMPONENT\_IDENTIFIER\}}.
\item \texttt{reasoning}: 1-3 sentences explaining why the hypothesis is consistent with the observation.
\end{itemize}
\end{tcolorbox}
\newpage
\subsection{Hypothesis Validation Stage}
\label{app:hypothesis-validation-prompt}
\begin{tcolorbox}[promptbox, title=Validation Plan Prompt]
You are currently in the \textbf{Hypothesis Validation} stage. Your goal is to design and execute a causal experiment that produces evidence either supporting or refuting the current hypothesis about \texttt{\{COMPONENT\_IDENTIFIER\}}. The outcome will determine whether this component's role is considered explained, or whether hypothesis generation restarts with revised constraints.

\textbf{Current hypothesis:} \texttt{\{CURRENT\_HYPOTHESIS\}}

\textbf{Prior observation:} \texttt{\{OBSERVATION\}}

\textbf{Requirement:} Design a causal experiment to test the hypothesis. Follow these instructions:
\begin{enumerate}[itemsep=2pt, parsep=0pt]
\item The experiment must be causal or interventional. Appropriate methods include activation patching, mean ablation, interchange intervention etc.
\item The plan must be directly motivated by the hypothesis. The expected result should follow logically from what the hypothesis predicts.
\item Be concrete and implementable. In the next step, Python code will be generated directly from this plan.
\item Keep each field concise, using 1-3 sentences.
\end{enumerate}

Respond with a JSON object containing exactly the following fields, and do not include any text outside the JSON:
\begin{itemize}[itemsep=2pt, parsep=0pt]
\item \texttt{stage}: \texttt{"VALIDATION PLAN for {COMPONENT\_IDENTIFIER}"}
\item \texttt{goal}: the specific prediction of the hypothesis being tested.
\item \texttt{procedure}: a short list of concrete experimental steps.
\item \texttt{inputs}: the model tensors, hooks, or cached activations required.
\item \texttt{expected\_result}: what should appear in the results if the hypothesis is correct.
\end{itemize}
\end{tcolorbox}

\begin{tcolorbox}[promptbox, title=Validation Code and Execution Prompt]
You are currently in the \textbf{Hypothesis Validation} stage, \textbf{Code Generation and Execution} step.

\textbf{Current hypothesis:} \texttt{\{CURRENT\_HYPOTHESIS\}}

\textbf{Validation plan:} \texttt{\{VALIDATION\_PLAN\}}

You have access to three tools:
\begin{itemize}[itemsep=2pt, parsep=0pt]
\item \tool{list\_directory(path)}: list files in a directory on the analysis server.
\item \tool{read\_file(path)}: read the full contents of a file on the analysis server.
\item \tool{execute\_python(code)}: execute Python code in the pre-configured analysis environment.
\end{itemize}

The execution environment has the following objects in scope:
\begin{itemize}[itemsep=2pt, parsep=0pt]
\item \texttt{model}: a \texttt{HookedTransformer}.
\item \texttt{tokenize(sequences)}: converts a list of token lists into a \texttt{[batch, seq\_len]} tensor. Each sequence starts with \texttt{BOS} and uses tokens from the task vocabulary.
\item \texttt{decode(logits)}: converts model output logits to task-semantic values in the same format as the task examples. The BOS position is excluded by default.
\item Standard imports.
\end{itemize}

Recommended steps:
\begin{enumerate}[itemsep=2pt, parsep=0pt]
\item Call \tool{list\_directory()} to discover available helper files with template code.
\item Read the file most relevant to the validation task to learn the available causal-intervention function signatures. Helper functions are already defined in the execution environment and should be called directly, without import statements.
\item Generate executable Python code for validating the hypothesis and call \tool{execute\_python}. The code must implement the validation plan, assign all outputs to a dictionary named \texttt{result}.
\end{enumerate}

Each \tool{execute\_python} call runs in a fresh subprocess; variables, imports, and state do not persist across calls. Every call must be self-contained. The experiment must include at least one causal intervention, such as ablation, activation patching, or interchange intervention. If a metric function is used, it should have signature \texttt{fn(logits: Tensor) -> float}, evaluate all output-bearing positions, exclude BOS, and use \texttt{decode()} to convert logits to task values. You have up to \texttt{\{MAX\_EXECUTE\_ROUNDS\}} execution attempts. If code returns an error, read the traceback, identify the root cause, and revise the next attempt.

\textbf{Decision criteria:}
\begin{itemize}[itemsep=2pt, parsep=0pt]
\item \texttt{support}: the results are consistent with the hypothesis, and the intervention effect aligns with the prediction.
\item \texttt{refute}: the results contradict the hypothesis, or the effect is absent, reversed, or inconsistent. If the result is ambiguous, prefer \texttt{refute}.
\end{itemize}

When sufficient experimental evidence has been gathered, respond with a JSON object containing the following fields, and do not include any text outside the JSON:
\begin{itemize}[itemsep=2pt, parsep=0pt]
\item \texttt{decision}: exactly \texttt{support} or \texttt{refute}.
\item \texttt{explanation}: 2--3 sentences summarizing the observed tool results and why they support the decision.
\item \texttt{ruling\_out}: required only for \texttt{refute}; specifies what claim in the hypothesis is contradicted.
\end{itemize}
\end{tcolorbox}
\subsection{Classification Stage}
\label{app:classification-prompt}
\begin{tcolorbox}[promptbox, title=Taxonomy Classification Prompt]
You are now in the \textbf{Taxonomy Classification} stage. All components have been analyzed. Assign each circuit component a functional role tag from the taxonomy below, based on its validated hypothesis.

\textbf{Taxonomy:}
\begin{itemize}[itemsep=2pt, parsep=0pt]
\item \texttt{INDICATOR}: Detects a property of a single input and emits a binary or predicate-like signal.
\item \texttt{AGGREGATOR}: Reduces content across multiple positions into a summary quantity, such as a count, fraction, or accumulated value. The information is collapsed after the operation.
\item \texttt{ROUTER}: Moves content from one position to another through positional or index-based selection. Content is copied across positions.
\item \texttt{MAPPER}: Transforms a single input at each position independently into a non-binary output, such as an arithmetic value, remapping, lookup, or reshaped representation, including when the result is a control signal.
\item \texttt{COMBINER}: Reads two or more distinct upstream signals or inputs and fuses them into one output.
\end{itemize}

\textbf{Validated hypotheses:} \texttt{\{VALIDATED\_HYPOTHESES\}}

\textbf{Instructions:}
\begin{enumerate}[itemsep=2pt, parsep=0pt]
\item Assign exactly one tag to each component based on the validated hypothesis and the taxonomy definitions.
\item When tags overlap, choose based on the component's functional input-output behavior.
\item Write a concise one-sentence description of what the component does in the context of this specific task.
\item Ground the description in the hypothesis, not in the generic taxonomy definition.
\item Do not output anything outside the JSON.
\end{enumerate}

Respond with a JSON object containing one entry for each component. Each entry should contain exactly two fields:
\begin{itemize}[itemsep=2pt, parsep=0pt]
\item \texttt{tag}: one of \texttt{INDICATOR}, \texttt{AGGREGATOR}, \texttt{ROUTER}, \texttt{MAPPER}, or \texttt{COMBINER}.
\item \texttt{description}: a one-sentence task-specific role description.
\end{itemize}
\end{tcolorbox}
\subsection{Summarization Stage}
\label{app:classification-prompt}
\begin{tcolorbox}[promptbox, title=Circuit Summarization Prompt]
You are in the \textbf{Circuit Summarization} stage. All components have been validated and classified.

\textbf{Localized components:} \texttt{\{LOCALIZED\}}

\textbf{Classified component roles:} \texttt{\{ROLES\}}

\textbf{Validated hypotheses:} \texttt{\{VALIDATED\_HYPOTHESES\}}

\textbf{Instructions:}
\begin{enumerate}[itemsep=2pt, parsep=0pt]
\item Synthesize a coherent account of how information flows through the localized circuit and what task is being implemented.
\item Describe the computational stages and the explicit interactions between components.
\item Infer the underlying task using both the input-output examples and the validated component mechanisms. If the mechanisms conflict with the examples, prioritize the examples.
\item Before finalizing the derived task description, check it against at least two input-output examples at non-boundary positions and revise if needed. Do not include these checks in the final output.
\item Write the derived task description as a concise task specification. Do not copy example values, or restate component-level operations.
\end{enumerate}

Respond with a valid JSON object containing the following fields, and do not include any text outside the JSON:
\begin{itemize}[itemsep=2pt, parsep=0pt]
\item \texttt{information\_flow}: 1-2 sentences describing the sequential dependencies among components.
\item \texttt{derived\_task\_description}: 1-2 sentences stating the sequential task performed by the model.
\end{itemize}
\end{tcolorbox}
\begin{table*}[t!]
\centering
\small
\setlength{\tabcolsep}{5pt}
\renewcommand{\arraystretch}{1.25}
\begin{tabularx}{\textwidth}{
    p{0.13\textwidth}
    p{0.04\textwidth}
    p{0.24\textwidth}
    X
}
\toprule
\textbf{Tag} & \textbf{Type} & \textbf{RASP primitive} & \textbf{Description} \\
\midrule
\textbf{INDICATOR} &
MLP &
\texttt{rasp.Map(pred, tokens)} & 
Detects a property of the current token and emits a binary signal. \\
\hline

\multirow{2}{*}{\textbf{AGGREGATOR}} &
\multirow{2}{*}{ATTN} &
\texttt{rasp.Aggregate()}, \texttt{rasp.SelectorWidth()} &
Computes a summary over selected positions (e.g.\ count, fraction, accumulated quantity). \\
\hline

\multirow{2}{*}{\textbf{ROUTER}} &
\multirow{2}{*}{ATTN} &
\texttt{rasp.Select(rasp.indices, \ldots)} + \texttt{rasp.Aggregate()} &
Moves a token from one position to another via positional or index-based selection. \\
\hline

\textbf{MAPPER} &
MLP &
\texttt{rasp.Map()} &
Applies an element-wise transformation to each position. \\
\hline

\multirow{2}{*}{\textbf{COMBINER}} &
\multirow{2}{*}{MLP} &
\texttt{rasp.SequenceMap()}, \texttt{rasp.LinearSequenceMap()} &
Reads and combines multiple upstream signals into one output through an arithmetic or logical operation. \\

\bottomrule
\end{tabularx}
\caption{Taxonomy of functional roles used in component-level annotations. Each tag captures the abstract computational role played by an attention head or MLP within a localized circuit, grounded in the corresponding RASP primitive.}
\label{tab:role_taxonomy}
\end{table*}

\section{Component Role Taxonomy}
\label{app:role-taxonomy}
Table~\ref{tab:role_taxonomy} contains details regarding the 5-class role taxonomy.

\section{Annotation Example: \texttt{frac\_prevs}}
\label{app:annotation-example}
This section illustrates how we derive component-level annotations from InterpBench using the \texttt{frac\_prevs} task as an example. The goal of \texttt{frac\_prevs} is to return, at each position, the fraction of previous tokens up to and including that position that are equal to \texttt{`x'}. Our annotation procedure uses two sources of information: the original RASP program, which specifies the high-level algorithm, and the high-level/low-level correspondence map, which identifies which trained InterpBench component implements each Tracr component. \\~\\
The RASP program for this task is:

\begin{verbatim}
is_x = (rasp.tokens == "x").named("is_x")
bools = rasp.numerical(is_x)
prevs = rasp.Select(rasp.indices,
                rasp.indices,
                rasp.Comparison.LEQ)
return rasp.numerical(
    rasp.Aggregate(prevs, bools, default=0)
).named("frac_prevs")
\end{verbatim}

This program decomposes the task into two main steps. First, \texttt{is\_x} computes a per-position predicate indicating whether the current token is \texttt{x}. The variable \texttt{bools} converts this predicate into a numerical signal. Second, \texttt{prevs} defines a prefix selector over positions, and \texttt{Aggregate(prevs, bools)} aggregates the \texttt{is\_x} signal over the prefix to compute the running fraction. Thus, \texttt{is\_x} corresponds to an indicator-style computation, while \texttt{frac\_prevs} corresponds to an aggregation over previous positions.

InterpBench provides a high-level/low-level correspondence map that links each Tracr high-level node to the trained low-level InterpBench component aligned with it. For \texttt{frac\_prevs}, the relevant entries are:

\begin{verbatim}
{TracrHLNode(
    name: blocks.0.mlp.hook_post,
    label: is_x_3,
    index: [:]
    ) : {
    LLNode(
        name='blocks.0.mlp.hook_post',
        index=[:])
    },


TracrHLNode(
    name: blocks.1.attn.hook_result,
    label: frac_prevs_1,
    index: [:, :, 0, :]
) : {LLNode(
    name='blocks.1.attn.hook_result',
    index=[:, :, 2, :])}}
is_x_3 |
HL = blocks.0.mlp.hook_post, index = [:]
-> LL = [('blocks.0.mlp.hook_post', [:])]
frac_prevs_1 |
HL = blocks.1.attn...,index=[:,:,0,:]
->  LL = [('blocks.1.attn...',[:,:,2,:])]
\end{verbatim}

The first correspondence entry maps the Tracr MLP component labeled \texttt{is\_x\_3} to the trained InterpBench component \texttt{blocks.0.mlp.hook\_post}. Since the corresponding RASP variable \texttt{is\_x} detects whether each token is \texttt{x}, we annotate this component as an \textsc{Indicator}. Its role note is: ``Computes a per-position feature indicating whether the token at that position is \texttt{x} or not.''

The second correspondence entry maps the Tracr attention output labeled \texttt{frac\_prevs\_1} to head 2 of \texttt{blocks.1.attn.hook\_result} in the trained InterpBench model. Since this component implements the aggregation over the prefix selector \texttt{prevs}, we annotate it as an \textsc{Aggregator}. Its role note is: ``Aggregates prefix fraction by attending over previous positions.'' We also record that this component uses the upstream \texttt{is\_x} feature computed by L0\_MLP.

The resulting component annotations are therefore:

\begin{verbatim}
components = [
    {
    "id": "L0_MLP",
    "hook": "blocks.0.mlp.hook_post",
    "role": {
        "tag": "INDICATOR",
        "note": "Computes per-position feature 
        indicating whether the token at that 
        position is 'x' or not."
    },
    "labels": ["is_x_3"],
    },
    {
    "id": "L1H2_ATTN",
    "hook": "blocks.1.attn.hook_result[2]",
    "role": {
        "tag": "AGGREGATOR",
        "note": "Aggregates prefix fraction 
        by attending over previous positions."
    },
    "labels": ["frac_prevs_1"],
    }
]
\end{verbatim}

This example shows how \benchmark extends InterpBench: InterpBench provides the trained low-level models and their correspondence to Tracr components, while \benchmark adds semantic role annotations by tracing each localized component back to the RASP variable it implements.

\section{\agent Walkthrough}
\label{app:agent-walkthrough}
To make the pipeline concrete, we trace \agent's full trajectory on component L0\_MLP for the running example \texttt{frac\_prevs}, using Claude-Sonnet-4.6 as the backbone.

\paragraph{Observation.} The observation plan is to characterize what L0\_MLP encodes, write code to cache its outputs across token types (\texttt{`x'}, \texttt{`c'}, \texttt{`a'}, \texttt{`b'}), and compare per-token-type difference vectors. \agent observes that L0\_MLP produces dramatically different outputs for \texttt{`x'} vs non-\texttt{`x'} tokens ($\lVert\Delta\rVert \approx 2.65$),
while non-\texttt{`x'} tokens are similar to each other
($\lVert\Delta\rVert \approx 0.04\text{-}0.07$).

\paragraph{Hypothesis.} \textit{``L0\_MLP is a binary \texttt{is\_x} feature detector: at every position, it writes a position-invariant signal into the residual stream encoding whether the token is \texttt{`x'} (positive) or not (near-zero/negative)''}

\paragraph{Validation.} \agent designs an activation-patching experiment: replace the L0\_MLP output for an \texttt{`x'} token with the output for a non-\texttt{`x'} token (and reverse). Patching confirms causal necessity, with normalized effect $\approx
0.97$ for $x \to c$ and $\approx 0.99$ on the reverse patch.
\paragraph{Classification.} \agent assigns the tag \textsc{Indicator}, matching the ground-truth annotation. It also writes a role description (\textit{``At each token position, L0\_MLP detects whether the token is `x' and writes a consistent binary feature into the residual stream''}) which closely matches the ground-truth note (\textit{``Computes
per-position feature indicating whether the token at that position is `x'.''}).
\paragraph{Summarization.} Based on the validated hypotheses and the assigned component tags, \agent defines the underlying task as: ``Given a sequence of tokens, the model outputs at each position the proportion of tokens seen so far (excluding \verb|<BOS>|) that are equal to `x', producing a running fraction that updates with each new token.''
\section{Evaluation Details and Human Annotation}
\label{app:eval-details}
\subsection{Overview}
Section~\ref{sec:eval-metrics} defines the main evaluation metrics. Here, we provide additional details about the human annotation protocol, agreement computation, process-level metrics, judge-bias analysis, run stability, and qualitative rubric examples. The human evaluation covers two final-output metrics, role description quality ($Q_{\text{desc}}$) and derived task accuracy ($Acc_{\text{task}}$), and two process-level metrics, validation-plan soundness ($S_{\text{val}}$) and hypothesis grounding.

The human evaluation was conducted in two stages. We first annotated outputs from the two backbones used in our initial analysis, GPT-5.4 and Claude-Sonnet-4.6. This larger GPT/Claude annotation set is used to report (i) the inter-annotator agreement and (ii) the agreement between human annotators and LLM judges in Table~\ref{tab:human-iaa-large}. It contains $n=110$ component-level instances for $Q_{\text{desc}}$ and $S_{\text{val}}$, and $n=62$ task-level instances for $Acc_{\text{task}}$.

After extending \agent to two additional backbones, Gemini-3.1-Pro and Qwen-3-Coder-30B-A3B-Instruct, we performed a second annotation pass on a smaller shared subset covering \textbf{all four backbones}. This cross-backbone subset contains 10 tasks and 17 components, yielding 68 component-level instances and 40 task-level instances. This subset supports the human-validation results and process-level comparisons discussed in the main text.

\subsection{Annotation Protocol}
\label{sec:human-eval-protocol}
For $Q_{\text{desc}}$, $Acc_{\text{task}}$, and $S_{\text{val}}$, the annotation was performed by two human annotators, both CS graduate students with machine-learning experience. The annotators were given a standardized annotation README, detailed metric definitions, and representative examples for each score level. The instructions followed the same rubrics used for the LLM-judge evaluation.

\par Annotators worked independently using a Streamlit-based interface. To reduce bias, model identities were hidden and randomized. The interface displayed model outputs using anonymized labels such as Agent A and Agent B; these labels were only interface labels and did not correspond to fixed backbone names. In the initial annotation stage, the interface showed outputs from GPT-5.4 and Claude-Sonnet-4.6. In the later cross-backbone annotation stage, the same blinding and randomization procedure was applied to outputs from all four backbones.

\par For each task, annotators first saw the task context, including the ground-truth task summary, up to five input-output examples, and the list of localized components with their ground-truth tags. For each localized component, annotators then saw the agent's hypothesis and validation plan as read-only context and rated validation-plan soundness ($S_{\text{val}}$) as \textit{Sound}, \textit{Partial}, or \textit{Unsound}. A \textit{Sound} plan directly tests the key mechanistic prediction of the hypothesis; a \textit{Partial} plan is causally relevant but indirect or incomplete; and an \textit{Unsound} plan does not meaningfully test the hypothesis.

\par Annotators next saw the ground-truth tag and role note for the component, followed by the agent's predicted role description. The predicted tag was shown only as context and was not itself rated. Annotators rated role description quality ($Q_{\text{desc}}$) as \textit{Correct}, \textit{Partial}, or \textit{Wrong}. A \textit{Correct} description captures the component's task-specific role; a \textit{Partial} description captures the main role but is vague, incomplete, or contains an incorrect mechanistic sub-claim; and a \textit{Wrong} description contradicts the reference role or describes a different function.

\par Finally, for each task, annotators saw the ground-truth task summary and each agent's derived task description. They rated task accuracy ($Acc_{\text{task}}$) as \textit{Correct} or \textit{Wrong}, indicating whether the derived description recovered the task-level behavior. Annotators could optionally provide a short rationale for each rating. The hidden mapping from anonymized agent labels to the underlying \agent backbone was recorded automatically for analysis but was not visible during annotation.

\subsection{Human Inter-Annotator Agreement Computation}
For ordinal 3-point metrics ($Q_{\text{desc}}$ and $S_{\text{val}}$), we report linearly weighted Cohen's $\kappa$. Linear weighting is appropriate because adjacent disagreements, such as 1 vs. 2, are less severe than endpoint disagreements, such as 0 vs. 2. For binary $Acc_{\text{task}}$, we report ordinary Cohen's $\kappa$ without weighting.
\par Table~\ref{tab:human-iaa-large} reports inter-annotator agreement on the larger GPT/Claude annotation subset. Agreement is \textbf{almost perfect} for the final-output metrics, with $\kappa=0.8$ for $Q_{\text{desc}}$ and $\kappa=0.96$ for $Acc_{\text{task}}$. Agreement is \textbf{moderate} for $S_{\text{val}}$ ($\kappa=0.46$), reflecting the greater subjectivity of judging whether a proposed causal experiment fully tests a mechanistic hypothesis. 
To rule out this subjectivity, we consider the lower score between the two annotators as the ground truth, implementing a stricter evaluation standard for LM agents. This applies to all human evaluations.

\subsection{Human-LLM Judge Agreement Computation}
We employ two LLM judges in our work. Similar to how we aggregate the annotated labels from the two annotators, we use conservative lower-score aggregation between the two LLM judges as well. That is, when we apply the LLM judges, we consider the lower score between them as the judging score for an agent. This aggregation retains a high score only when both annotators or both LLM judges assign it, reducing the chance of over-crediting an incomplete or incorrect explanation. We report the human-LLM judge agreement in Table~\ref{tab:human-iaa-large}, the bottom panel, where \textit{lower human} label is the lower of the two human annotator scores, and the \textit{lower judge} label is the lower of the two LLM-judge scores.
% We use conservative lower-score aggregation when comparing human and LLM-judged labels. For each example, the \textit{lower human} label is the lower of the two human annotator scores, and the \textit{lower judge} label is the lower of the two LLM-judge scores. This aggregation retains a high score only when both annotators or both LLM judges assign it, reducing the chance of over-crediting an incomplete or incorrect explanation. 
\begin{table}[h]
\centering
\small
\begin{tabular}{lcc}
\toprule
Metric & $n$ & $\kappa$ \\
\midrule
\multicolumn{3}{l}{\textbf{Human--human agreement}} \\
\midrule
$Q_{\text{desc}}$ & 110 & 0.802 \\
$S_{\text{val}}$  & 110 & 0.460 \\
$Acc_{\text{task}}$ & 62 & 0.963 \\
\midrule
\multicolumn{3}{l}{\textbf{Lower human vs. Lower judge agreement}} \\
\midrule
$Q_{\text{desc}}$ & 110 & 0.753 \\
$S_{\text{val}}$  & 110 & 0.481 \\
$Acc_{\text{task}}$ & 62 & 0.864 \\
\bottomrule
\end{tabular}
\caption{
Human--human inter-annotator agreement and Human--LLM-judge agreement on the larger GPT/Claude annotation subset. The value of $n$ counts model-output instances, $\kappa$ denotes linearly weighted Cohen's $\kappa$ for ordinal metrics.
}
\label{tab:human-iaa-large}
\end{table}
% \par Table~\ref{tab:human-iaa-large} reports Cohen's $\kappa$ between these two aggregated labels. 
Table~\ref{tab:human-iaa-large} shows that the agreement is \textbf{substantial} for $Q_{\text{desc}}$ ($\kappa=0.75$) and \textbf{almost perfect} for $Acc_{\text{task}}$ ($\kappa=0.86$), supporting the use of LLM judges for the final-output metrics. In contrast, agreement is lower for $S_{\text{val}}$ ($\kappa=0.48$). Together with the lower human-human agreement for $S_{\text{val}}$, this suggests that validation-plan soundness is useful as a process-level diagnostic in qualitative analysis but less reliable as an LLM-judged headline metric. We therefore opt not to use it as an official metric for \benchmark and leave more reliable automatic evaluation of validation-plan quality to future work.

\subsection{Cross-Backbone Human Validation Subset}
We also evaluate a smaller subset covering all four \agent backbones. This subset contains 10 tasks and 17 localized components, yielding 68 component-level instances for $Q_{\text{desc}}$ and $S_{\text{val}}$, and 40 task-level instances for $Acc_{\text{task}}$.
Table~\ref{tab:human-iaa-cross} reports the corresponding agreement results on the cross-backbone subset covering all four \agent backbones. The final output metrics show high human--human and human--LLM agreement, while $S_{\text{val}}$ remains lower, supporting our decision to treat it as a process-level metric.
\begin{table}[h]
\centering
\small
\begin{tabular}{lcc}
\toprule
Metric & $n$ & $\kappa$ \\
\midrule
\multicolumn{3}{l}{\textbf{Human--human agreement}} \\
\midrule
$Q_{\text{desc}}$ & 68 & 0.83 \\
$S_{\text{val}}$  & 68 & 0.37 \\
$Acc_{\text{task}}$ & 40 & 0.96 \\
\midrule
\multicolumn{3}{l}{\textbf{Lower human vs. Lower judge agreement}} \\
\midrule
$Q_{\text{desc}}$ & 68 & 0.76 \\
$S_{\text{val}}$  & 68 & 0.44 \\
$Acc_{\text{task}}$ & 40 & 0.80 \\
\bottomrule
\end{tabular}
\caption{
Human--human inter-annotator agreement and Human--LLM-judge agreement on the cross-backbone human-validation subset covering all four \agent backbones. The value of $n$ counts model-output instances, $\kappa$ denotes linearly weighted Cohen's $\kappa$ for ordinal metrics.
}
\label{tab:human-iaa-cross}
\end{table}

For completeness, Table~\ref{tab:per-judge-iaa} reports pairwise agreement between each human annotator and each LLM judge on the cross-backbone subset. Agreement varies across individual judge pairs, especially for $S_{\text{val}}$, but remains higher for the two final-output metrics.

\begin{table}[h]
\centering
\small
\begin{tabular}{l c c c c}
\toprule
Metric & H1-GPT & H1-Gemini & H2-GPT & H2-Gemini \\
\midrule
$Q_{\text{desc}}$  & 0.73 & 0.78 & 0.63 & 0.74 \\
$S_{\text{val}}$   & 0.23 & 0.59 & 0.43 & 0.49 \\
$Acc_{\text{task}}$& 0.73 & 0.74 & 0.73 & 0.74 \\
\bottomrule
\end{tabular}
\caption{Linear-weighted Cohen's $\kappa$ between each LLM judge and each human annotator (H1, H2) on the \textbf{cross-backbone subset} ($n=68$ for $Q_{\text{desc}}/S_{\text{val}}$, $n=40$ for
$Acc_{\text{task}}$).}
\label{tab:per-judge-iaa}
\end{table}

{
\subsection{Judge Bias Analysis}
\label{app:judge-bias}
We report per-judge scores for the LLM-judged metrics to examine whether judges favor outputs from their own model family. Table~\ref{tab:per-judge-results} shows that both judges produce the same backbone ranking for $Acc_{\text{task}}$. For $Q_{\text{desc}}$, both rank GPT-5.4 third and Qwen fourth, while the near-tied Claude and Gemini backbones exchange the top two positions, but with minimal difference in scores. 

We find no consistent advantage for a judge's own model family. The Gemini judge is generally more lenient on $Q_{\text{desc}}$, including toward GPT-generated descriptions, while the GPT judge assigns lower scores across most backbones. Each judge gives its own backbone a slightly higher $Acc_{\text{task}}$, but the overall ranking remains unchanged. Our reported score takes the minimum of the two judgments for each example, preventing either judge from unilaterally inflating the final score. Together with the blinded human evaluation in Section~\ref{sec:human-eval-protocol}, these results suggest that our main comparisons are not driven by the choice of LLM judge.
\begin{table}[h]
\centering
\small
\setlength{\tabcolsep}{3pt}
\begin{tabular}{llcccc}
\toprule
Metric & Judge & GPT-5.4 & Claude & Gemini & Qwen \\
\midrule
$Q_{\text{desc}}$
& GPT-5.4 & 0.48 & 0.59 & 0.60 & 0.28 \\
& Gemini-3.1 & 0.58 & 0.68 & 0.67 & 0.27 \\
\midrule
$Acc_{\text{task}}$
& GPT-5.4 & 0.73 & 0.77 & 0.85 & 0.25 \\
& Gemini-3.1 & 0.67 & 0.77 & 0.90 & 0.29 \\
\bottomrule
\end{tabular}
\caption{Per-judge results for the LLM-judged metrics. Columns correspond to the \agent backbone, and rows report scores assigned by each judge.}
\label{tab:per-judge-results}
\end{table}

\subsection{Run Stability}
\label{app:run-stability}
We run \agent three times per backbone on a fixed subset of 31 tasks containing 55 components. The subset was selected to approximately preserve the component distribution of the full benchmark. Since this analysis uses a subset, the mean scores are not expected to match the full-benchmark results exactly. We use the same decoding configuration as the main evaluation, temperature 0 for the closed-source backbones and 0.7 for Qwen, with all other settings left at their provider defaults.

Table~\ref{tab:run-stability} reports the mean and standard deviation across the three runs. Stability varies across backbones and metrics. GPT-5.4 shows the greatest variation, primarily in $S_{\text{exec}}$, while Qwen is comparatively stable despite its lower overall performance. The main final-output trends remain stable. Gemini achieves the highest $Q_{\text{desc}}$ and $Acc_{\text{task}}$, Claude achieves the highest $Acc_{\text{tag}}$, and Qwen remains weakest on $Q_{\text{desc}}$ and $Acc_{\text{task}}$.
\begin{table}[h]
\centering
\small
\setlength{\tabcolsep}{2.5pt}
\begin{tabular}{lcccc}
\toprule
Metric & GPT-5.4 & Claude-4.6 & Gemini-3.1 & Qwen-30B \\
\midrule
$Acc_{\text{tag}}$
& $0.74 \pm 0.02$
& $0.78 \pm 0.01$
& $0.76 \pm 0.01$
& $0.75 \pm 0.01$ \\
$Q_{\text{desc}}$
& $0.46 \pm 0.01$
& $0.56 \pm 0.01$
& $0.57 \pm 0.03$
& $0.27 \pm 0.01$ \\
$Acc_{\text{task}}$
& $0.58 \pm 0.04$
& $0.72 \pm 0.05$
& $0.86 \pm 0.01$
& $0.29 \pm 0.00$ \\
$S_{\text{exec}}$
& $0.52 \pm 0.07$
& $0.95 \pm 0.01$
& $0.64 \pm 0.00$
& $0.43 \pm 0.03$ \\
\bottomrule
\end{tabular}
\caption{Mean and standard deviation across three runs on a fixed subset of 31 tasks containing 55 components. Higher is better.}
\label{tab:run-stability}
\end{table}

}

\subsection{Process-Level Diagnostics}
In addition to the final-output metrics, we analyze two process-level diagnostics: hypothesis grounding and validation-plan soundness. These diagnostics help identify where the agent succeeds or fails inside the observe-hypothesize-validate loop.
\subsubsection{Hypothesis Grounding}
We annotate hypothesis grounding on the same 10-task, 17-component cross-backbone subset used for the main-text human validation. This annotation was performed by one author of the paper for analysis purposes.
% by one human annotator, a CS graduate student with machine-learning experience. 
For each \agent backbone and each component, the annotator was shown the natural-language observation produced by the agent, the subsequent hypothesis generated from that observation, and the task context, including the task description, input-output examples, and ground-truth component roles. The annotator judged whether the hypothesis was supported by the observation on a 3-point scale: 0 if the hypothesis contradicted or ignored the observation, 1 if it was partially supported but added unsupported details, and 2 if it was fully supported by the observation. 

Table~\ref{tab:hyp-ground-result} summarizes the grounding scores. The proprietary backbones produce mostly observation-grounded hypotheses, while Qwen-3-Coder more often adds unsupported task-specific details beyond its observations.

\begin{table}[h]
\centering
\small
\begin{tabular}{lcccc}
\toprule
Backbone & Mean & Fully grounded & Partial \\
\midrule
GPT-5.4 & 1.94 & 94.1\% & 5.9\% \\
Claude-Sonnet-4.6 & 1.94 & 94.1\% & 5.9\% \\
Gemini-3.1-Pro & 1.94 & 94.1\% & 5.9\% \\
Qwen-3-Coder-30B & 1.41 & 41.2\% & 58.8\% \\
\bottomrule
\end{tabular}
\caption{
Human-evaluated hypothesis-grounding results on the 10-task, 17-component cross-backbone subset. The score measures whether the agent's hypothesis is supported by its own observation (scale: 0-2).
}
\label{tab:hyp-ground-result}
\end{table}
\subsubsection{Validation-Plan Soundness}
Validation-plan soundness ($S_{\text{val}}$) measures whether a proposed validation experiment directly tests the current hypothesis. A sound plan should specify an intervention whose expected result follows from the hypothesis and whose outcome could meaningfully support or refute it. We report LLM-judged $S_{\text{val}}$ in Table~\ref{tab:sval} as a process-level diagnostic for analyzing validation behavior.
However, as shown in Tables~\ref{tab:human-iaa-large},~\ref{tab:human-iaa-cross}, and ~\ref{tab:per-judge-iaa}, $S_{\text{val}}$ has lower human--human and human--LLM agreement than the final-output metrics. This suggests that validation-plan soundness is more subjective to evaluate than role descriptions or task descriptions. For this reason, the main text reports validation-plan soundness using human annotations on the cross-backbone subset. The LLM-judged scores in Table~\ref{tab:sval} are included only as an additional diagnostic. The LLM-judged scores show a broadly consistent backbone-level pattern.
\begin{table}[h]
\centering
\small
\begin{tabular}{l c }
\toprule
Backbone & $S_\text{val}$  \\
\midrule
GPT-5.4 & 1.85 \\
Claude-Sonnet-4.6 & 1.0 \\
Gemini-3.1-Pro & 1.13 \\
Qwen-3-Coder-30B & 0.76 \\
\bottomrule
\end{tabular}
\caption{LLM-judged validation-plan soundness for the four \agent backbones. Scores are averaged over all validation plans using the conservative lower score assigned by the two LLM judges, GPT-5.4 and Gemini-3.1-Pro.}
\label{tab:sval}
\end{table}

\subsection{Qualitative Rubric Examples}
\subsubsection{Role Description Quality}
\label{app:qdesc-ex}
We provide examples for the three $Q_{\text{desc}}$ score levels. As described in Section~\ref{sec:eval-metrics}, $Q_{\text{desc}}$ evaluates whether the predicted role description matches the reference component role. We use a 3-point scale when prompting the LLM judge and later rescale it to $[0, 1]$ for ease of reporting. A score of 2 indicates a correct task-specific description, a score of 1 indicates a partially correct description with missing or incorrect mechanistic details, and a score of 0 indicates a wrong or contradictory description.

\paragraph{Correct example (score 2): L0\_MLP in \texttt{frac\_prevs}}
\begin{itemize}
\item
  \textbf{Reference tag:} INDICATOR \\ 
  \textbf{Reference note:} Computes
  per-position feature indicating whether the token at that position is `x' or not.
\item
  \textbf{Agent's description:} At each token position, L0\_MLP detects whether the token is `x' and writes a consistent binary feature vector into the residual stream encoding `is\_x' (positive) or `not-x' (near-zero/negative).
\item
  \textbf{Rationale:} The description
  accurately captures that L0\_MLP computes a binary indicator of
  whether each token is `x' and writing this to the residual stream.
\end{itemize}

\paragraph{Partial example (score 1): L1H2 of \texttt{frac\_prevs}}
\begin{itemize}
\item
  \textbf{Reference tag:} AGGREGATOR \\
  \textbf{Reference note:} Aggregates prefix fraction by attending over previous positions.
\item \textbf{Agent's description:} At each position $t$, L1H2 reads the `is\_x' feature vectors from all prior positions, attends preferentially to `x' positions, and collapses them
  into a single output vector encoding the running fraction of
  `x' tokens seen so far $(count\_x / t)$.
\item \textbf{Rationale:} The description correctly identifies L1H2 as aggregating information from previous positions to compute a running fraction of `x' tokens. However, it incorrectly says the head attends preferentially to x positions.  The ground-truth attention pattern is uniform over previous positions, not preferential toward `x' positions, and the fraction is computed by averaging the upstream `is\_x' values.
\end{itemize}

\paragraph{Wrong example (score 0): L1H0 from an ``extract-unique tokens'' task.}

\begin{itemize}
\item
  \textbf{Reference tag:} AGGREGATOR \\ 
  \textbf{Reference note:} Aggregates matching positions, defined by same token and earlier-or-equal index, into a per-position count of how many times each token has appeared up to and including the current position.
\item
  \textbf{Agent's description:} L1H0 mainly
  routes residual content by preserving local state through
  self-attention on `c' positions and otherwise sometimes pulling a weak, largely non-essential generic contextual write from a recent `c'-associated position.
\item
  \textbf{Rationale:} The description focuses on preserving local state and attending to `c' positions, which does not match the ground-truth role of aggregating same-token prefix positions to compute occurrence counts.
\end{itemize}

\subsubsection{Hypothesis Grounding}
Hypothesis grounding evaluates whether \agent's hypothesis follows from its own observation. This score is separate from role-description correctness. A hypothesis can be grounded in the observation but still be wrong with respect to the reference role, or correct in outcome but unsupported by the evidence the agent cites.

\noindent\textbf{Fully grounded example (score 2).}
In the \texttt{frac\_prevs} task, \agent observes that the L2 norm of L0\_MLP outputs varies between \texttt{`x'} and non-\texttt{`x'} tokens. It then hypothesizes that L0\_MLP is a binary \texttt{`is\_x'} feature detector. 

This receives a score of 2 because the hypothesis is \textbf{directly supported} by the observation. The observed activation difference is exactly the kind of evidence expected from a token-property indicator.

\noindent\textbf{Partially grounded example (score 1).}
In the spam-keyword detection task, \agent observes that L0\_MLP has high, stable activation norms across positions dominated by a small set of neurons. It then hypothesizes that the component detects position-specific spam patterns by aggregating features from previous tokens. 

This receives a score of 1. Although the observation \textbf{supports} the broad claim that L0\_MLP is important and neuron-mediated, it \textbf{does not support} the more specific claims about position-specific behavior or aggregation over previous tokens.

\noindent\textbf{Ungrounded example (score 0).}
In the sequence-length multiplication task, \agent observes that L0\_MLP has high activation norms with position-dependent top neurons. It then hypothesizes that L0\_MLP applies a non-linear transformation to each token. 

This receives a score of 0. The hypothesis \textbf{does not follow} from the observation: position-dependent activation strength does not provide evidence for a token-wise non-linear transformation.

\subsubsection{Validation-Plan Soundness}
\label{app:sval}
$S_{\text{val}}$ evaluates whether the agent's proposed validation experiment directly tests its stated hypothesis. The score does not judge whether the hypothesis itself is correct, nor whether the generated code eventually executes successfully. Instead, it asks whether the proposed causal experiment would meaningfully support or refute the specific mechanistic claim made in the hypothesis.

We use a 3-point scale:
\begin{itemize}
\item \textbf{Sound (2):} The plan directly targets the key prediction in the hypothesis. The intervention cleanly distinguishes the hypothesis from nearby alternatives.
\item \textbf{Partial (1):} The plan is causally motivated and relevant, but it tests the hypothesis only indirectly, bundles multiple subclaims together, or leaves important alternatives unresolved.
\item \textbf{Unsound (0):} The plan does not test the stated hypothesis. For example, it may test a different claim, rely only on non-causal evidence, or propose an expected result that would actually refute the hypothesis.
\end{itemize}
\noindent\textbf{Partial example (score 1): L1H2 in \texttt{frac\_prevs}.}
\begin{itemize}
\item \textbf{Task:} The model computes the running fraction of \texttt{x} tokens seen so far.
\item \textbf{Component:} L1H2, whose reference role is to aggregate prefix information by attending over previous positions.
\item \textbf{Agent hypothesis:} L1H2 is a running-fraction aggregator. It reads upstream \texttt{is\_x} features from L0\_MLP, attends to prior positions, and writes an output vector encoding the running fraction of \texttt{x} tokens.
\item \textbf{Validation plan:} Run the model on sequences with different running fractions, mean-ablate L1H2, and measure how the clean-minus-ablated output changes with the running fraction.
\end{itemize}

The plan is relevant because it performs a causal intervention on L1H2. If ablating this head systematically disrupts the running-fraction output, that would provide evidence that the head contributes to the task. Thus, the plan tests the general necessity of L1H2 for the running-fraction computation.

However, the plan is incomplete because the hypothesis makes more specific mechanistic claims than simple necessity. It claims that L1H2 reads \texttt{is\_x} features from previous positions and encodes the running fraction. Mean ablation alone does not distinguish whether the head uniformly aggregates previous positions, attends preferentially to \texttt{x} positions, or contributes through another nearby aggregation strategy. We therefore rate this plan as Partial: it tests the right general mechanism, but it does not cleanly isolate the key prediction in the hypothesis.

\section{API Cost}
\label{app:api-cost}
We estimate the API cost of running \agent on the full 84-case \benchmark benchmark. We count tokens using each provider's native tokenizer API (Claude, Gemini) and \texttt{tiktoken} for GPT-5.4 and Qwen. Claude-Sonnet-4.6 incurs the largest estimated API cost, at \$147.55 total (\$1.76 per task), followed by GPT-5.4 at \$77.47 total (\$0.92 per task) and Gemini-3.1-Pro at \$33.81 total (\$0.40 per task). Qwen-3-Coder is self-hosted, so we report \$0 marginal API cost and exclude GPU-hour costs; however, the full run required approximately 10 GPU-hours, which we exclude from the dollar-cost estimate because GPU cost depends on the hardware and pricing environment.

The cost differences highlight the cost-performance tradeoff across backbones. Claude produces the highest $Acc_{tag}$, $Q_{desc}$, and $S_{exec}$, but is also the most expensive. Gemini achieves the best $Acc_{task}$ while being $4\times$ cheaper than Claude, and GPT-5.4 falls between them with the highest $S_\text{val}$. 
We report the token statistics and API Cost in Table~\ref{tab:api-cost}.
\begin{table*}[t]
\centering
\small
\begin{tabular}{lrrrr}
\toprule
\textbf{Backbone} & \textbf{Input tokens} & \textbf{Output tokens} & \textbf{Total cost} & \textbf{Mean cost/case} \\
\midrule
Claude-Sonnet-4.6 & 34.49M & 2.94M & \$147.55 & \$1.76 \\
GPT-5.4 & 16.73M & 2.38M & \$77.47 & \$0.92 \\
Gemini-3.1-Pro & 12.77M & 0.69M & \$33.81 & \$0.40 \\
Qwen-3-Coder-30B-A3B & 18.64M & 1.31M & \$0.00 & -- \\
\bottomrule
\end{tabular}
\caption{
Estimated token usage and API cost for running \agent on the full 84-case \benchmark benchmark. Qwen-3-Coder-30B-A3B is self-hosted, so we report zero marginal API cost and exclude GPU-hour costs.
}
\label{tab:api-cost}
\end{table*}

\section{Real Circuit Reference Annotation}
\label{sec:appendix-realckt}
\begin{table*}[!t]
    \centering
    \small
 \setlength{\tabcolsep}{5pt}
    \renewcommand{\arraystretch}{1.25}
    \begin{tabularx}{\textwidth}{
        p{0.06\textwidth}
        p{0.18\textwidth}
        p{0.14\textwidth}
        X
    }
    \toprule
    \textbf{Comp.} & \textbf{Reference Role} & \textbf{Observation} & \textbf{Hypothesis loop} \\
    \midrule

    L15H13
    &
    \textbf{Primary $B$-transfer head.} Reads $B$ at the final query position, writes $B$-dependent information into the residual stream.
    &
    Final token attention concentrates strongly on operand $B$, near-zero mass elsewhere. \newline
    &
    \textbf{\textit{Hypothesis} (Iteration 1 \textcolor{darkgreen}{\checkmark{}}):} L15H13 routes $B$'s identity to the final token. Ablation should cause a significant accuracy drop, with errors clustering near $A{+}C$. \newline
    \textbf{\textit{Validation}:} Supported. Ablation yields a large accuracy drop, activation patching restores it. \\
    \hline

    L16H1
    &
    \textbf{Tertiary $C$-transfer head.} Invisible in the full model but load-bearing once the stronger $C$ routes (L15H3, L15H31) are suppressed.
    &
    Final-token attention to operand $C$, with secondary attention to \verb|<BOS>| and smaller weights on $B$ and $A$
    &
    \textbf{\textit{Hypothesis} (Iteration 1 \textcolor{darkred}{\ding{55}{}}):} L16H1 is a \emph{primary} $C$-router. Ablation should drop accuracy significantly. \newline
    \textbf{\textit{Validation}:} Refuted. Ablation yields no accuracy drop.
    \par\smallskip\hrule\smallskip
    \textbf{\textit{Hypothesis} (Iteration 2 \textcolor{darkgreen}{\checkmark{}}):} L16H1 is a \emph{backup} $C$-router. Invisible under solo ablation, active when L15H3 and L15H31 are removed. \newline
    \textbf{\textit{Validation}:} Supported. Ablation together with (L15H3, L15H31) causes a significant further accuracy drop. \\

    \bottomrule
    \end{tabularx}
    \caption{Per-component exploration trace for two transfer heads of the AF1 circuit on Llama-3-8B. \agent converges immediately on L15H13's role as the primary $B$-transfer head, but requires a refuted iteration before re-hypothesizing L16H1 as a backup $C$-router.}
    \label{tab:agent-exploration}
\end{table*}
\subsection{Setup}
\label{sec:realckt-setup}
\paragraph{Task and model.} We use the three-operand addition prompt template ``$A + B + C = \quad $'' with $A, B, C \in \{0, 1, \dots , 100\}$ and the answer lying in the range $\{0, 999\}$, evaluated on Llama-3-8B. Our case study builds on the All-for-One (AF1) circuit, which identifies a sparse subgraph sufficient for this arithmetic behavior.
\paragraph{Localized Circuit.} 
Starting from the AF1 subgraph, we construct a 10-component localized circuit for explanation. The circuit contains
five transfer attention heads in layers 15 and 16 (L15H3, L15H13, L15H31, L16H1, L16H21), three late MLPs (L20\_MLP, L29\_MLP, L31\_MLP), two late attention heads with strong logit-lens signal (L26H3, L28H18). We deliberately retain some causally redundant components (L29\_MLP, L31\_MLP, L26H3, L28H18) to test whether \agent can distinguish mechanistic evidence from suggestive but non-causal signals.

\paragraph{Reference annotation.}
AF1 establishes the high-level arithmetic circuit, but it does not provide the component-level roles needed for our evaluation.
We therefore construct manual reference
annotations for the 10 localized components. We start
from the AF1 subgraph and run targeted interventions
on 99 prompts of the form ``$A + B + C = \quad$'',
restricted to examples the model answers correctly.
The raw model has baseline accuracy $1.00$ on this set.

For attention heads, we inspect final-token attention patterns, edge ablations, last-query head ablations, and corrupt-operand activation patching. For MLPs, we use zero/mean/CAMA-style ablations, corrupt-operand patching, iterative pruning, and logit-lens projections. These experiments distinguish primary operand-transfer
heads, backup transfer heads, a late MLP that is necessary
for accuracy but does not directly write in the answer
direction, and components with strong logit-lens signal
but weak causal effect. Tables~\ref{tab:af1-transfer-annotations} and~\ref{tab:af1-late-annotations} summarize the resulting reference roles.
\newcolumntype{Y}{>{\raggedright\arraybackslash}X}
\begin{table*}[t]
\centering
\small
\setlength{\tabcolsep}{4pt}
\begin{tabularx}{\textwidth}{p{0.1\textwidth} p{0.24\textwidth} Y}
\toprule
\textbf{Component} & \textbf{Reference role} & \textbf{Evidence used for annotation} \\
\midrule
L16H21 &
\textbf{Primary $A$-transfer head}. This head carries information about the first operand to the final token. &
At the final token, L16H21 almost always attends to operand $A$. Mean attention on $A$ is $0.979$, and $A$ is the top key in all $99$ prompts. This attention is causally important because zeroing this head's final-token output reduces accuracy from $1.000$ to $0.192$. Removing only the $A$ edge reduces accuracy from $1.000$ to $0.101$, while removing the $B$ or $C$ edge has no effect. In a corrupt-$A$ patch, the model switches to the
corrupt answer on $98.0\%$ of prompts, and the corrupt-vs-clean answer margin moves from $-4.750$ to $+4.156$. \\
\midrule
L15H13 &
\textbf{Primary $B$-transfer head}. This head moves information about operand $B$ to the final token. &
  At the final token, L15H13 puts most of its attention on $B$ (mean mass $0.878$, top key in $98/99$ prompts). The $B$ edge was also found to be causally relevant. Zeroing the head's final-token output drops accuracy from $1.000$ to $0.374$, and removing only the $B$ edge drops it to $0.465$. But removing $A$ or C operand edges has no effect. Corrupt-$B$ patching changes the corrupt-answer rate from $0.000$ to $0.475$ and gives a corrupt-answer logit gain of $
  +2.953$. \\
\midrule
L15H3 &
\textbf{Primary $C$-transfer head.} This head moves information about operand $C$ to the final token, but its effect is weaker than the primary $A$ and $B$ movers because $C$ has \textbf{backup transfer routes}. &
At the final token, L15H3 puts most of its attention on $C$ (mean mass $0.843$, top key in $99/99$ prompts). Zeroing the head's final-token output drops accuracy from $1.000$ to $0.747$. The $C$ edge is the causal
  one as removing only the $C$ edge drops accuracy to $0.879$, while removing \verb|<BOS>|, $A$, $B$, or \texttt{=} has no effect. Corrupt-$C$ patching gives a corrupt-answer logit gain of $+0.734$, but only changes the corrupt-answer rate from $0.000$ to $0.020$. In iterative pruning over L15-L16 heads, L15H3 is the final survivor; removing it takes accuracy from $0.030$ to $0.000$. \\
\midrule
L15H31 &
\textbf{Backup $C$-transfer head.} This head carries $C$-related information, but it is mostly redundant while L15H3 is active. Suppressing L15H3 exposes L15H31 as a load-bearing backup $C$ route. &
At the final token, L15H31 attends mostly to $C$, with some attention to $B$ (mean mass $0.505$ on C and $0.340$ on $B$; top key $C$ in $90/99$ prompts). In the full model, zeroing its final-token output only drops
accuracy from $1.000$ to $0.980$. After suppressing L15H3, zeroing L15H31 drops accuracy from $0.747$ to $0.414$, and removing only its $C$ edge gives the same accuracy. This shows that L15H31's $C$ edge becomes
  important when L15H3 is absent. Corrupt-$C$ patching in this setting gives a corrupt-answer logit gain of $+0.762$, compared with only $+0.221$ in the full model. \\
\midrule
L16H1 &
\textbf{Tertiary $C$-transfer backup head.} The head carries $C$-related signal, but it becomes cleanly load-bearing only after the stronger $C$-transfer routes L15H3 and L15H31 are both suppressed. &
At the final token, L16H1 attends mostly to $C$, with substantial attention to \verb|<BOS>| (mean mass $0.516$ on $C$ and $0.219$ on \verb|<BOS>|). With only L15H3 suppressed, removing the $C$ edge already hurts
accuracy, from $0.747$ to $0.515$. After suppressing both L15H3 and L15H31, zeroing L16H1 drops accuracy from $0.414$ to $0.121$, and removing only its $C$ edge gives the same
accuracy. In this double-suppressed setting, corrupt-$C$ patching gives a corrupt-answer logit gain of $+0.875$. \\
\bottomrule
\end{tabularx}
\caption{
Manual reference annotations for the AF1 transfer heads. The reference roles distinguish primary operand-transfer heads from backup C-transfer heads.
}
\label{tab:af1-transfer-annotations}
\end{table*}

\begin{table*}[t]
\centering
\small
\setlength{\tabcolsep}{4pt}
\begin{tabularx}{\textwidth}{p{0.1\textwidth} p{0.2\textwidth} Y}
\toprule
\textbf{Component} & \textbf{Reference role} & \textbf{Evidence used for annotation} \\
\midrule
L20\_MLP &
\textbf{Latent arithmetic feature builder}. L20 is useful for the arithmetic task, but its output does not look like a direct answer vector or a clean intermediate such as $A+B$, $A+C$, or $B+C$. &
Zeroing L20\_MLP drops accuracy from $1.000$ to $0.737$. CAMA-style ablation gives a smaller but nonzero drop of $0.192$. Corrupt-operand patching gives similar corrupt-answer flip rates for $A$, $B$, and $C$ ($0.253$, $0.242$, $0.273$), with corrupt-answer logit gains of $+2.03$, $+2.20$, and $+2.31$, suggesting that L20 does not strongly prefer one operand over the others. Also, candidate-target logit lens is weak. The answer top-5 rate is only $0.051$, and pair-sum targets remain near zero. Directionally, the output is only weakly aligned with the answer direction ($\cos=0.036$) and is not an amplification of the pre-MLP residual ($\cos=-0.188$). We therefore annotate L20 as a latent feature builder rather than an explicit answer writer. \\
\midrule
L29\_MLP &
\textbf{Answer-related but redundant MLP}. L29's output points toward the correct answer in projection tests, but removing it does not hurt the model on this task. &
A candidate-target logit lens on L29\_MLP recovers the answer at top-5 rate $0.394$, with mean answer logit $+8.35$, while pair-sum and operand targets stay near zero. Direction decomposition gives
DLA(answer) $=+8.35$ compared with DLA(random) $=+0.97$, so the output is answer-related. However, direct zero-ablation leaves accuracy unchanged ($1.000$ to $1.000$), and CAMA-style ablation also gives zero
drop. Corrupt-operand patching gives nonzero corrupt-answer logit gains around $+1.3$, but never flips the prediction. We therefore annotate L29 as answer-related but causally redundant in the full circuit. \\
\midrule
L31\_MLP &
\textbf{Strong answer projection but causally redundant MLP}. L31 has the strongest answer signal under logit-lens-style projection, but the signal is broad rather than answer-specific, and the component is not
necessary in isolation. &
This MLP has the strongest single-MLP answer lens signal, with answer top-5 rate $0.566$ and mean answer logit $+15.67$. Direction decomposition also gives large DLA(answer) $=+15.67$ compared with DLA(random) $=-0.16$. However, the projection is not specific to the final answer: pair sums and individual operands also receive high mean logits ($A+B$: $14.57$, $B+C$: $14.54$, $A+C$: $14.48$, $A$ $+15.07$, $B$ $+14.96$, $C$ $+14.98$). Isolated zero-ablation barely changes accuracy ($1.000 \rightarrow 0.990$), CAMA-style ablation gives zero drop, and corrupt-operand patching produces zero corrupt-answer flips. We therefore annotate L31 as lens-positive but causally redundant in the full circuit. \\
\midrule
L26H3 &
\textbf{Lens-positive \verb|<BOS>|-sink head}. This late attention head has answer-related projection under logit lens, but its attention is concentrated on \verb|<BOS>| rather than the operand tokens. &
Per-head logit lens ranks L26H3 second among late attention heads, with top-1 rate $0.162$ and top-3 rate $0.394$. Its final-token attention is dominated by \verb|<BOS>|. \verb|<BOS>| mass is $0.587$, while total operand mass is
only $0.027$ ($21.7\times$ smaller). In targeted pruning over the top late-attention lens heads, removing L26H3 leaves accuracy at $1.000$. We therefore annotate it as a lens-positive \verb|<BOS>|-sink head rather than a
causally necessary arithmetic component. \\
\midrule
L28H18 &
\textbf{Lens-positive \verb|<BOS>| / anchor head}. This is the strongest late-attention logit-lens head, but its attention is not primarily on operands and it is causally redundant in isolation. &
Per-head logit lens ranks L28H18 first among late attention heads, with top-1 rate $0.222$ and top-3 rate $0.455$. Its final-token attention is \verb|<BOS>|-leaning: \verb|<BOS>| mass is $0.445$, while total operand mass is
$0.102$ ($4.4\times$ smaller). The remaining non-\verb|<BOS>| mass is concentrated more on positional anchors such as \texttt{=} than on operands. In targeted pruning over the top late-attention lens heads, removing
L28H18 leaves accuracy at $1.000$. We therefore annotate it as lens-positive but causally redundant. \\
\bottomrule
\end{tabularx}
\caption{
  Manual reference annotations for the late-layer AF1 components. These components distinguish causal task-relevant computation from answer-correlated projection evidence. L20\_MLP is causally relevant
  and carries operand-dependent signal, but does not directly write the final answer or a clean pair-sum representation. In contrast, L29\_MLP, L31\_MLP, L26H3, and
  L28H18 show answer-related projection signals but have little or no isolated causal effect in the full circuit.
  }
\label{tab:af1-late-annotations}
\end{table*}

% \newpage

\newpage
\section{Artifact Use, Licensing, and Data Content}
\label{app:artifact-use}
This work uses existing research artifacts and software libraries for MI evaluation. \benchmark is built on InterpBench and Tracr-derived models, which we use as controlled research testbeds with known circuit structure. We use TransformerLens and LangGraph as implementation libraries for model inspection and agent orchestration. We also use Llama-3-8B and Qwen-3-Coder-30B-A3B-Instruct only for research evaluation and do not redistribute third-party model weights. Our released artifacts, including benchmark annotations, prompt templates, and \agent code, are intended for research use in circuit-explanation evaluation and should not be interpreted as deployment-ready guarantees of model safety. We release our code and annotations under the MIT License. Our data are based on synthetic algorithmic tasks derived from InterpBench/RASP programs, consisting of task tokens and program outputs rather than human-authored or user-provided text. We manually checked the task vocabularies, input-output examples, prompt templates, annotations, and agent outputs for human names, uniquely identifying information, and offensive content. The Llama-3-8B case study uses arithmetic prompts only. We found no PII or offensive content requiring anonymization.

\end{document}